 \documentclass[tablecaption=bottom,wcp]{jmlr} 

\usepackage{booktabs}
\usepackage[load-configurations=version-1]{siunitx} 

\theorembodyfont{\upshape}
\theoremheaderfont{\scshape}
\theorempostheader{:}
\theoremsep{\newline}

\usepackage{xspace}
\usepackage{graphicx}
\usepackage{booktabs} 
\usepackage{adjustbox}
\usepackage{siunitx}
\usepackage{bm}
\usepackage[hashEnumerators,smartEllipses,hybrid]{markdown}

\newcommand{\ie}{i.e.\ }
\newcommand{\eg}{e.g.\ }
\newcommand{\egnows}{e.g.} 
\newcommand{\wrt}{w.r.t.\ }
\newcommand{\iid}{i.i.d.\ }

\newcommand{\ours}{second-moment loss\ }
\newcommand{\oursnows}{second-moment loss} 

\newcommand{\btheta}{\theta}
\newcommand{\bthetaTilde}{\tilde{\theta}}
\newcommand{\f}{f_{\btheta}}
\newcommand{\fsub}{f_{\bthetaTilde}}

\newcommand{\erf}{\operatorname{Erf}}

\newcommand{\mmc}{\mu_\text{drop}}
\newcommand{\smc}{\sigma_\text{drop}}

\newif\ifdraft
\usepackage{color}

\drafttrue 

\ifdraft
\usepackage{cancel}
\fi

\jmlrproceedings{AABI 2021}{3rd Symposium on Advances in Approximate Bayesian Inference, 2021}

\title[Non-parametric Regression Uncertainty]{A Novel Regression Loss for Non-Parametric \mbox{Uncertainty Optimization}}

  \author{\Name{Joachim Sicking} \Email{joachim.sicking@iais.fraunhofer.de}\\
   \Name{Maram Akila} \Email{maram.akila@iais.fraunhofer.de}\\
   \Name{Maximilian Pintz} \Email{maximilian.alexander.pintz@iais.fraunhofer.de}\\
   \Name{Tim Wirtz} \Email{tim.wirtz@iais.fraunhofer.de}\\
   \addr Fraunhofer IAIS, Sankt Augustin, Germany
    \AND
  \Name{Asja Fischer} \Email{asja.fischer@uni-bochum.de}\\
  \addr University of Bochum, Bochum, Germany
    \AND
  \Name{Stefan Wrobel} \Email{stefan.wrobel@cs.uni-bonn.de}\\
  \addr Fraunhofer IAIS, Sankt Augustin, Germany\\
  \addr Fraunhofer Center for Machine Learning\\
  \addr University of Bonn, Bonn, Germany}

\begin{document}

\maketitle

\begin{abstract}
Quantification of uncertainty is one of the most promising  approaches to establish \textit{safe} machine learning. Despite its importance, it is far from being generally solved, especially for neural networks. One of the most commonly used approaches so far is Monte Carlo dropout, which is computationally cheap and easy to apply in practice. However, it can underestimate the uncertainty. We propose a new objective, referred to as second-moment loss (SML), to address this issue. While the full network is encouraged to model the mean, the dropout networks are explicitly used to optimize the model variance. We intensively study the performance of the new objective on various UCI regression datasets. Comparing to the state-of-the-art of deep ensembles, SML leads to comparable prediction accuracies and uncertainty estimates while only requiring a single model. Under distribution shift, we observe moderate improvements. As a side result, we introduce an intuitive Wasserstein distance-based uncertainty measure that is non-saturating and thus allows to resolve quality differences between any two uncertainty estimates.
\end{abstract}

\section{Introduction}
\label{sec:intro}

Having attracted great attention in both academia and digital economy, deep neural networks (DNNs, \citet{goodfellow2016deep}) are about to become vital components of safety-critical applications. Examples are autonomous driving \citep{pomerleau1989alvinn,bojarski2016end} or medical diagnostics \citep{liu2014early}, where prediction errors potentially put humans at risk. 
These systems require methods that are robust not only under lab conditions (\iid data sampling), but also under continuous domain shifts, think \eg of adults on \mbox{e-scooters} or growing varieties of mobile health sensors. Besides shifts in the data, the data distribution itself poses further challenges. Critical situations are (fortunately) rare and thus strongly under-represented in datasets. Despite their rareness, these critical situations have a significant impact on the safety of operations. This calls for comprehensive self-assessment capabilities of DNNs and recent uncertainty mechanisms can be seen as a step in that direction. 

\begin{figure}[bth]
    \centering
    \includegraphics[width=0.7\textwidth]{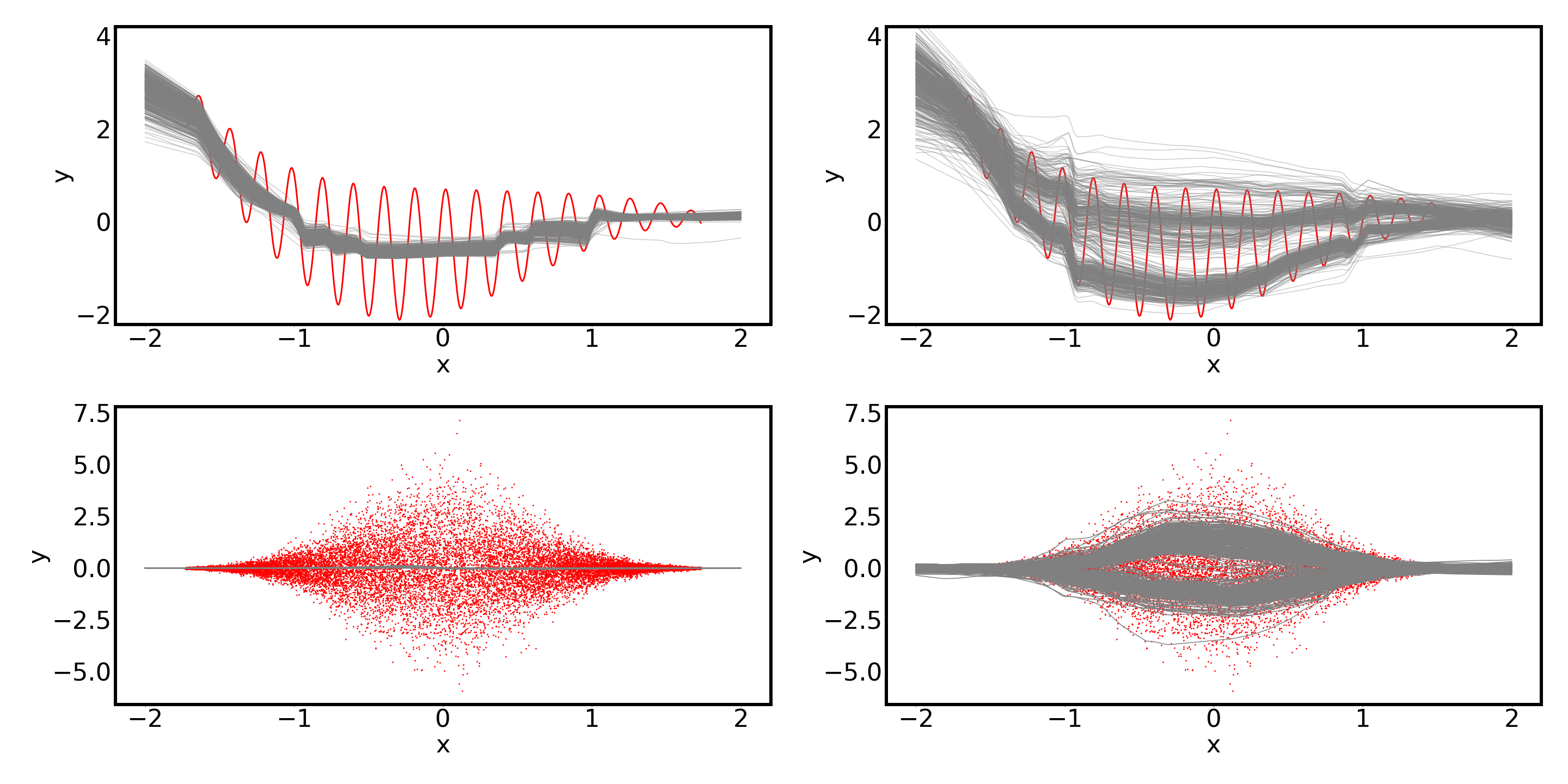}
    \caption{Sampling-based uncertainty mechanisms on toy datasets. The \ours (right) induces uncertainties that capture aleatoric uncertainty. This is in contrast to MC dropout (left). Ground truth data is shown in red. Each grey line represents the outputs of one of 200 sub-networks that are obtained by applying dropout-based sampling to the trained full network.}
    \label{fig:subnets_toydata}
\end{figure}

While a variety of uncertainty approaches have been established, stable quantification of uncertainty is still an open problem. Many recent machine learning applications are \mbox{\eg equipped} with Monte Carlo (MC) dropout \citep{gal2016dropout} that offers conceptual simplicity and scalability. 
However, it tends to underestimate uncertainties thus bearing disadvantages compared to more recent approaches such as deep ensembles \citep{lakshminarayanan2017simple}. We propose an alternative uncertainty mechanism. It builds on dropout sub-networks and explicitly optimizes variances (see Fig.\ \ref{fig:subnets_toydata} for an illustrative example). Technically, this is realized by a simple additive loss term, the \textit{\oursnows}. To address the above outlined requirements for safety-critical systems, we evaluate our approach systematically \wrt continuous data shifts.

In detail, our contribution is the introduction of a novel regression loss  for better calibrated uncertainties applicable to dropout networks, reaching state-of-the-art performance in an empirical study and improving on it when considering data shifts.


\section{Related work}
\label{sec:relWork}

Approaches to estimate predictive uncertainties can be broadly categorized into three groups: Bayesian approximations, ensemble approaches and parametric models.  

Monte Carlo dropout \citep{gal2016dropout} and its variants (see \eg \citet{gal2017concrete,kendall2017uncertainties,postels2019sampling}) are prominent representatives of the first group. They are theoretically well understood (see \eg \cite{sicking2020characteristics}), offer a Bayesian motivation, conceptual simplicity and scalability to application-size neural networks (NNs). This combination distinguishes MC dropout from other Bayesian neural network (BNN) approximations like \citet{blundell2015weight} and \citet{ritter2018scalable}. Note that dropout training is also used---independent from an uncertainty context---for better model generalization \citep{srivastava2014dropout}.

Ensembles of neural networks, so-called deep ensembles \citep{lakshminarayanan2017simple}, pose another popular approach to uncertainty modelling. Comparative studies of uncertainty mechanisms \citep{snoek2019can,gustafsson2020evaluating} highlight their advantageous uncertainty quality, making deep ensembles a state-of-the-art method. \citet{fort2019deep} argue that deep ensembles capture multi-modality of loss landscapes and thus yield potentially more diverse sets of solutions.

The third group are parametric modelling approaches that extend point estimations by adding a model output that is interpreted as variance or covariance~\citep{nix1994estimating,heskes1997practical}.
Typically, these approaches optimize a (Gaussian) negative log-likelihood (NLL, \citet{nix1994estimating}). A more recent representative of this group is, \egnows, \cite{kendall2017uncertainties}, for a review see \cite{khosravi2011comprehensive}. A closely related model class is deep kernel learning  combining NNs and Gaussian processes (GPs) in various ways (see \eg \citep{wilson2016deep,iwata2017improving,garnelo2018neural,qiu2019quantifying}).

The quality of uncertainties is typically evaluated using negative log-likelihood ~\citep{blei2006variational,walker2016uncertain,gal2016dropout}, expected calibration error (ECE) ~\citep{naeini2015obtaining, snoek2019can}, its variants, and by considering relations between uncertainty estimates and model errors \citep{workshop_unc_realism}.


\section{Second-moment loss}
\label{sec:ours}

Monte Carlo (MC) dropout was proposed as a computationally cheap approximation of performing Bayesian inference in neural networks \citep{gal2016dropout}. Given a  neural network \mbox{$\f : \mathbb{R}^d\to \mathbb{R}^m$} with parameters $\btheta$, MC dropout samples sub-networks $\fsub$ by randomly dropping nodes from the main model $\f$. During MC dropout inference the prediction is given by the mean estimate over the predictions of a given sample of sub-networks, while the uncertainty associated with this prediction can be estimated, e.g., in terms of the sample variance. During MC dropout training the objective function, in our case the mean squared error (MSE), is applied to the sub-networks separately. Due to this training procedure, all sub-network predictions are shifted towards the same training targets, which can result in overconfident predictions, \ie in an underestimation of prediction uncertainty.\footnote{An intuitive explanation is as follows: Let $\f$ be a NN with one-dimensional output. For MC dropout with the MSE loss we get $\left\langle(\fsub(x) - y)^2\right\rangle = (\left\langle\fsub(x)\right\rangle - y)^2 + \sigma^2(\fsub(x)) $. Therefore, it simultaneously minimizes the squared error between sub-network mean and target and the variance $\sigma^2(\fsub(x)) = \langle \fsub^2(x) \rangle - \langle \fsub(x) \rangle^2$ over the sub-networks.}

Based on this observation, we propose to use the sub-networks $\fsub$ in a different way: they are explicitly \textit{not} encouraged to fit the data mean directly. This is the task of the full network $\f$. The sub-networks $\fsub$ instead model aleatoric uncertainty and prediction residuals if the prediction of the full network $\f$ is incorrect.  Thus, we deliberately assign different `jobs' to the main network $\f$ on the one hand and its sub-networks on the other hand. Formalizing this idea into an optimization objective yields
\begin{equation}\label{eq:mod_mc_objective_train}
    L = L_{\rm regr} + L_{\rm sml} = \frac{1}{M}\,\sum_{i=1}^M \Big[\underbrace{{\left({\f(x_i)} - y_i \right)}^2}_{\rm regression\ loss} +\ \beta\,\underbrace{(\,\lvert{\fsub(x_i)} - \f(x_i) \rvert - \lvert \f(x_i) - y_i \rvert)^2}_{\rm second-moment \, loss} \,\Big] \enspace,
\end{equation}
where the sum runs over a mini-batch of size  $M\!<\!N$ taken from the set of observed samples $\mathcal{D} = \{(x_i,y_{i})\}_{i=1}^N$,  $x_i\in\mathbb{R}^d$ denotes the input, $y_{i}\in\mathbb{R}^m$ the ground-truth label, and $\beta>0$ is a hyper-parameter that weights both terms. The first term, $L_{\rm regr}$, is the MSE \wrt the full network $\f$. The second term, $L_{\rm sml}$, seeks to optimize\footnote{To avoid unintended optimization of full $\f$ in direction of $\fsub$, we only back-propagate through $\fsub$ in $L_{\rm sml}$.} the sub-networks $\fsub$. It aims at finding sub-networks such that the distance $|\fsub - \f|$ matches the aleatoric uncertainty or the prediction residual which is quantified by $| \f(x_i) - y_i |$.\footnote{As our choice of $L_{\rm sml}$ removes all directional information of the residual, possible (optimal) solutions for the $\fsub$ are not uniquely determined. For a one-dimensional example based on aleatoric uncertainty see appendix \ref{appendix:analytical}.}
This leads to a significant increase in the variance of the sub-networks, \ie the second moment of $\fsub$, compared to standard MC dropout, which is why we name $L_\text{sml}$ the \textit{\ours} (SML).\footnote{For brevity, we also refer to the entire loss objective $L$ as \ours during evaluation.} 
The standard deviations $\sigma_{\rm total}$ of the predictions of the sub-networks \wrt the prediction of the mean network induced by the SML have two components: the spread $\sigma_{\rm drop}$ of the sub-networks and an offset $\left\vert \f - \langle \fsub\rangle \right\vert$ between the full network and the sub-network mean that our loss might cause, concretely, $\sigma_\text{total}=\smc + \vert \f - \langle \fsub\rangle \vert$. While $\vert \f - \langle \fsub\rangle \vert$ is 
reminiscent of residual matching, $\smc$ seems to be more closely related to modelling uncertainties. We show in appendix \ref{appendix:sigma_total_eval} that $\smc$ accounts on average for more than $80\%$ of $\sigma_{\rm total}$ in our experiments.

Note that while we investigate the proposed objective in terms of dropout sub-networks in this paper, our arguments as well as the actual approach are generally applicable to other models that allow to formulate sub-networks given some kind of mean model. Besides the regression tasks considered here our approach could be useful for other objectives which use or benefit from an underlying distribution, \eg uncertainty quantification in classification.


\section{Experiments}
\label{sec:experiments}
\label{sec:uci}

We study uncertainty quality on UCI regression datasets, where we extend the dataset selection in \citet{gal2016dropout} by adding three further datasets: `diabetes', `california', and `superconduct'. 
Apart from \iid train- and test-data results, we study regression performance and uncertainty quality \textit{under data shift}. Such distributional changes and uncertainty quantification are closely linked since the latter ones are rudimentary ``self-assessment'' mechanisms that help to judge model reliability. These judgements gain importance for model inputs that are \textit{structurally different} from train data. Appendix \ref{appendix:uci_eval} elaborates on our ways of splitting the data, namely \textit{pca-based} splits in input space (using the first principal component) and \textit{label-based} splits. 
We assess uncertainty performance in terms of the expected calibration error (ECE) and Wasserstein distance (WS) and regression performance using root-mean-squared error (RMSE) and negative log-likelihood (NLL). All measures are described in detail in appendix \ref{appendix:experimentalSetup}, where you can also find more details on the network, the implementation of the methods and the training procedure.
For brevity of exposition, we limit our discussion here largely to the ECE. 
An evaluation of the other measures can be found in appendix \ref{appendix:uci_eval}.
All presented results are $5$- or $10$-fold cross validated.

\begin{figure}[bth]
    \centering
    \includegraphics[width=1.0\textwidth]{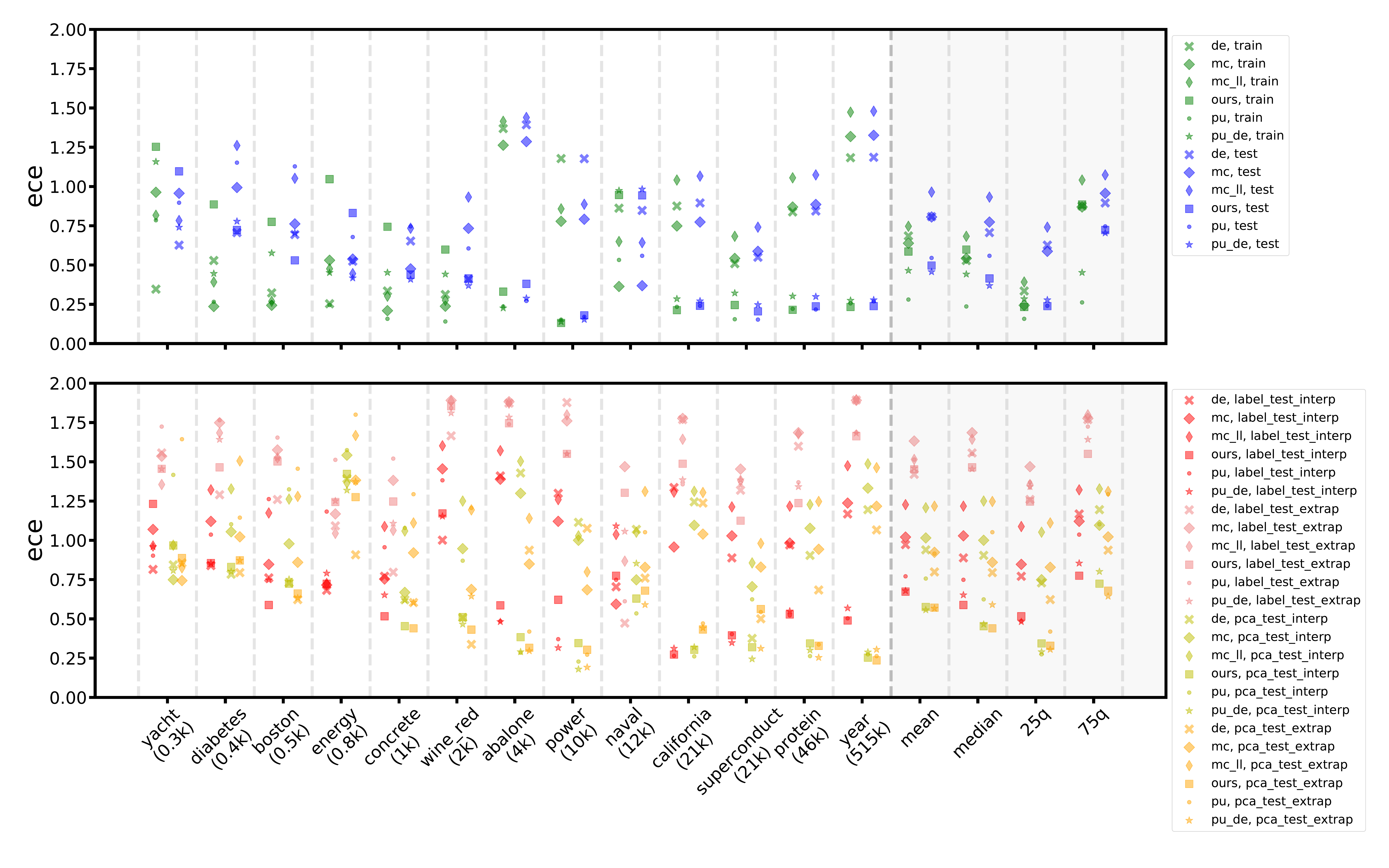}
    \vspace{-3em}
    \caption{Expected calibration errors (ECEs) for 13 UCI regression datasets under \iid conditions (top) and under data shift (bottom). Uncertainty methods are encoded via plot marker, data splits via color. Each plot point corresponds to a cross-validated trained network. Summarizing statistics (rhs) are indicated by a light grey background.}
    \label{fig:uci_ece}
\end{figure}

Fig.\ \ref{fig:uci_ece} provides ECEs for 13 UCI datasets that are sorted by dataset size on the \mbox{x-axis}. The top panel shows train- (green) and test-set (blue) ECEs, the bottom panel test-set ECEs under two pca-based data shifts (yellow-green, orange) and two label-based data splits (red, light red), for inter- and extrapolation respectively. Uncertainty methods are encoded via plot markers, \eg PU-DE as `star' and SML-trained networks (`ours') as `square'. We summarize these dataset-specific results on the right hand side of the figure (light grey background). The columns `mean' and `median' of this summary show that on training sets, ECEs are smallest for PU, followed by PU-DE and the SML network. On test data, however, PU, PU-DE and the SML network share the first place.
Looking at the stability \wrt data shift, \ie the ability to extra- or interpolate to ``unseen'' data, PU loses in performance while PU-DE and SML reach the smallest calibration errors in three out of four cases, compare the lower panel in Fig.\ \ref{fig:uci_ece}.

Summarizing these evaluations, we find SML to be as strong as the state-of-the-art method of PU-DEs while using only a single network compared to an ensemble of $5$ networks. We moreover observe advantages for SML under PCA- and label-based data shifts. Three datasets lead to overestimated uncertainties for the SML, see discussion in appendix \ref{appendix:uci_eval}. A visual tool to further inspect uncertainty quality are residual-uncertainty scatter plots as shown in appendix \ref{appendix:res_error}. For a reflection on NLL and comparisons of the different uncertainty measures see appendix \ref{appendix:uci_eval}.

\section{Conclusion}
\label{sec:disc}

We approach dropout-based uncertainty quantification from a new direction: sub-networks are explicitly not encouraged to model the data mean, they capture aleatoric uncertainties and potential fitting residuals of the full network instead. Technically, this is realized by an additional loss term that accompanies the standard regression objective: the \textit{\oursnows}. Our loss enables stable training. Training complexity and runtime behavior at inference are comparable to MC dropout. Task performances and uncertainty qualities of these models are on par with (parametric) deep ensembles, the widely used state-of-the-art for uncertainty quantification. However, unlike deep ensembles, we use single networks. In practice, this might allow to reduce training effort significantly compared to deep ensembles, especially for application-scale networks. Moreover, a single network requires only a fraction of the storage of a deep ensemble, making models with competitive uncertainties more accessible for mobile or embedded applications.

An extensive study of uncertainties under data shift revealed advantages of SML-trained models compared to deep ensembles: while both methods \textit{on average} provide comparable results, we find a higher stability across a variety of datasets and data shifts for the SML. 
Technically, we attribute this gain in stability to our sub-network-based approach: like MC dropout, we integrate uncertainty estimates into the very structure of the network, rendering it more robust towards unseen inputs than a parameter estimate.

Moreover, the \ours can serve as a drop-in replacement for MC dropout on regression tasks. For already trained MC dropout models, post-training with the \ours might suffice to improve uncertainty quality. As an outlook, our first such post-training experiments are encouraging.
Another interesting variant is the combination of SML with last-layer dropout (MC-LL) as it enables sampling-free inference \citep{postels2019sampling}. Preliminary experiments show clearly improved uncertainty qualities compared to standard MC-LL. A potentially interesting avenue for near real-time applications.

The simple additive structure of the \ours makes it applicable to a variety of optimization objectives. For classification, we might be able to construct a non-parametric counterpart to prior networks \citep{malinin2018predictive}.
Taking a step back, we demonstrated an easily feasible approach to influence and train sub-network distributions.
This could be a promising avenue, for distribution matching but also for theoretical investigations.

\acks{The research of J.\ Sicking and M.\ Akila was funded by the German Federal Ministry for Economic Affairs and Energy within the project ``KI Absicherung – Safe AI for Automated Driving''. Said authors would like to thank the consortium for the successful cooperation. The work of T.\ Wirtz was funded by the German Federal Ministry of Education and Research, ML2R - no. 01S18038B. S.\ Wrobel contributed as part of the Fraunhofer Center for Machine Learning within the Fraunhofer Cluster for Cognitive Internet Technologies. The work of A.~Fischer was supported by the Deutsche Forschungsgemeinschaft (DFG, German Research Foundation) under Germany’s Excellence Strategy – EXC-2092 \textsc{CaSa} – 390781972.}

\bibliography{references}

\begin{thebibliography}{38}
\providecommand{\natexlab}[1]{#1}
\providecommand{\url}[1]{\texttt{#1}}
\expandafter\ifx\csname urlstyle\endcsname\relax
  \providecommand{\doi}[1]{doi: #1}\else
  \providecommand{\doi}{doi: \begingroup \urlstyle{rm}\Url}\fi

\bibitem[Arjovsky et~al.(2017)Arjovsky, Chintala, and
  Bottou]{arjovsky2017wasserstein}
Martin Arjovsky, Soumith Chintala, and L{\'e}on Bottou.
\newblock Wasserstein {GAN}.
\newblock \emph{arXiv preprint arXiv:1701.07875}, 2017.

\bibitem[Bishop(2006)]{bishop2006pattern}
Christopher~M Bishop.
\newblock \emph{Pattern recognition and machine learning}.
\newblock Springer, 2006.

\bibitem[Blei et~al.(2006)Blei, Jordan, et~al.]{blei2006variational}
David~M Blei, Michael~I Jordan, et~al.
\newblock Variational inference for {D}irichlet process mixtures.
\newblock \emph{Bayesian Analysis}, 1\penalty0 (1):\penalty0 121--143, 2006.

\bibitem[Blundell et~al.(2015)Blundell, Cornebise, Kavukcuoglu, and
  Wierstra]{blundell2015weight}
Charles Blundell, Julien Cornebise, Koray Kavukcuoglu, and Daan Wierstra.
\newblock Weight uncertainty in neural networks.
\newblock \emph{arXiv preprint arXiv:1505.05424}, 2015.

\bibitem[Bojarski et~al.(2016)Bojarski, Del~Testa, Dworakowski, Firner, Flepp,
  Goyal, Jackel, Monfort, Muller, Zhang, et~al.]{bojarski2016end}
Mariusz Bojarski, Davide Del~Testa, Daniel Dworakowski, Bernhard Firner, Beat
  Flepp, Prasoon Goyal, Lawrence~D Jackel, Mathew Monfort, Urs Muller, Jiakai
  Zhang, et~al.
\newblock End to end learning for self-driving cars.
\newblock \emph{arXiv preprint arXiv:1604.07316}, 2016.

\bibitem[Foong et~al.(2019)Foong, Li, Hern{\'a}ndez-Lobato, and
  Turner]{foong2019between}
Andrew~YK Foong, Yingzhen Li, Jos{\'e}~Miguel Hern{\'a}ndez-Lobato, and
  Richard~E Turner.
\newblock `{I}n-between' uncertainty in {B}ayesian neural networks.
\newblock \emph{arXiv preprint arXiv:1906.11537}, 2019.

\bibitem[Fort et~al.(2019)Fort, Hu, and Lakshminarayanan]{fort2019deep}
Stanislav Fort, Huiyi Hu, and Balaji Lakshminarayanan.
\newblock Deep ensembles: A loss landscape perspective.
\newblock \emph{arXiv preprint arXiv:1912.02757}, 2019.

\bibitem[Gal and Ghahramani(2016)]{gal2016dropout}
Yarin Gal and Zoubin Ghahramani.
\newblock Dropout as a {B}ayesian approximation: Representing model uncertainty
  in deep learning.
\newblock In \emph{International Conference on Machine Learning}, pages
  1050--1059, 2016.

\bibitem[Gal et~al.(2017)Gal, Hron, and Kendall]{gal2017concrete}
Yarin Gal, Jiri Hron, and Alex Kendall.
\newblock Concrete dropout.
\newblock In \emph{Advances in Neural Information Processing Systems}, pages
  3581--3590, 2017.

\bibitem[Garnelo et~al.(2018)Garnelo, Schwarz, Rosenbaum, Viola, Rezende,
  Eslami, and Teh]{garnelo2018neural}
Marta Garnelo, Jonathan Schwarz, Dan Rosenbaum, Fabio Viola, Danilo~J Rezende,
  SM~Eslami, and Yee~Whye Teh.
\newblock Neural processes.
\newblock \emph{arXiv preprint arXiv:1807.01622}, 2018.

\bibitem[Glorot et~al.(2011)Glorot, Bordes, and Bengio]{relu3}
Xavier Glorot, Antoine Bordes, and Yoshua Bengio.
\newblock Deep sparse rectifier neural networks.
\newblock volume~15 of \emph{Proceedings of Machine Learning Research}, pages
  315--323, Fort Lauderdale, FL, USA, 11--13 Apr 2011. JMLR Workshop and
  Conference Proceedings.

\bibitem[Gneiting and Raftery(2007)]{gneiting2007strictly}
Tilmann Gneiting and Adrian~E Raftery.
\newblock Strictly proper scoring rules, prediction, and estimation.
\newblock \emph{Journal of the American Statistical Association}, 102\penalty0
  (477):\penalty0 359--378, 2007.

\bibitem[Goodfellow et~al.(2016)Goodfellow, Bengio, and
  Courville]{goodfellow2016deep}
Ian Goodfellow, Yoshua Bengio, and Aaron Courville.
\newblock \emph{Deep learning}.
\newblock MIT press, 2016.

\bibitem[Gustafsson et~al.(2020)Gustafsson, Danelljan, and
  Schon]{gustafsson2020evaluating}
Fredrik~K Gustafsson, Martin Danelljan, and Thomas~B Schon.
\newblock Evaluating scalable {B}ayesian deep learning methods for robust
  computer vision.
\newblock In \emph{Proceedings of the IEEE/CVF Conference on Computer Vision
  and Pattern Recognition Workshops}, pages 318--319, 2020.

\bibitem[Heskes(1997)]{heskes1997practical}
Tom Heskes.
\newblock Practical confidence and prediction intervals.
\newblock In \emph{Advances in Neural Information Processing Systems}, pages
  176--182, 1997.

\bibitem[Iwata and Ghahramani(2017)]{iwata2017improving}
Tomoharu Iwata and Zoubin Ghahramani.
\newblock Improving output uncertainty estimation and generalization in deep
  learning via neural network {G}aussian processes.
\newblock \emph{arXiv preprint arXiv:1707.05922}, 2017.

\bibitem[Kendall and Gal(2017)]{kendall2017uncertainties}
Alex Kendall and Yarin Gal.
\newblock What uncertainties do we need in {B}ayesian deep learning for
  computer vision?
\newblock In \emph{Advances in Neural Information Processing Systems}, pages
  5574--5584, 2017.

\bibitem[Khosravi et~al.(2011)Khosravi, Nahavandi, Creighton, and
  Atiya]{khosravi2011comprehensive}
Abbas Khosravi, Saeid Nahavandi, Doug Creighton, and Amir~F Atiya.
\newblock Comprehensive review of neural network-based prediction intervals and
  new advances.
\newblock \emph{IEEE Transactions on Neural Networks}, 22\penalty0
  (9):\penalty0 1341--1356, 2011.

\bibitem[Kingma and Ba()]{kingma2014adam}
Diederik~P Kingma and Jimmy Ba.
\newblock Adam: A method for stochastic optimization.
\newblock \emph{ICLR 2015}.

\bibitem[Kuleshov et~al.(2018)Kuleshov, Fenner, and Ermon]{Kuleshov2018}
Volodymyr Kuleshov, Nathan Fenner, and Stefano Ermon.
\newblock {Accurate uncertainties for deep learning using calibrated
  regression}.
\newblock In \emph{35th International Conference on Machine Learning, ICML
  2018}, 2018.

\bibitem[Lakshminarayanan et~al.(2017)Lakshminarayanan, Pritzel, and
  Blundell]{lakshminarayanan2017simple}
Balaji Lakshminarayanan, Alexander Pritzel, and Charles Blundell.
\newblock Simple and scalable predictive uncertainty estimation using deep
  ensembles.
\newblock In \emph{Advances in Neural Information Processing Systems}, pages
  6402--6413, 2017.

\bibitem[Liu et~al.(2014)Liu, Liu, Cai, Pujol, Kikinis, and Feng]{liu2014early}
Siqi Liu, Sidong Liu, Weidong Cai, Sonia Pujol, Ron Kikinis, and Dagan Feng.
\newblock Early diagnosis of {A}lzheimer's disease with deep learning.
\newblock In \emph{2014 IEEE 11th International Symposium on Biomedical Imaging
  (ISBI)}, pages 1015--1018. IEEE, 2014.

\bibitem[Malinin and Gales(2018)]{malinin2018predictive}
Andrey Malinin and Mark Gales.
\newblock Predictive uncertainty estimation via prior networks.
\newblock In \emph{Advances in Neural Information Processing Systems}, pages
  7047--7058, 2018.

\bibitem[Naeini et~al.(2015)Naeini, Cooper, and
  Hauskrecht]{naeini2015obtaining}
Mahdi~Pakdaman Naeini, Gregory~F Cooper, and Milos Hauskrecht.
\newblock Obtaining well calibrated probabilities using {B}ayesian binning.
\newblock In \emph{Proceedings of the AAAI Conference on Artificial
  Intelligence. AAAI Conference on Artificial Intelligence}, volume 2015, page
  2901, 2015.

\bibitem[Nix and Weigend(1994)]{nix1994estimating}
David~A Nix and Andreas~S Weigend.
\newblock Estimating the mean and variance of the target probability
  distribution.
\newblock In \emph{Proceedings of IEEE International Conference on Neural
  Networks 1994}, volume~1, pages 55--60. IEEE, 1994.

\bibitem[Pomerleau(1989)]{pomerleau1989alvinn}
Dean~A Pomerleau.
\newblock Alvinn: An autonomous land vehicle in a neural network.
\newblock In \emph{Advances in Neural Information Processing Systems}, pages
  305--313, 1989.

\bibitem[Postels et~al.(2019)Postels, Ferroni, Coskun, Navab, and
  Tombari]{postels2019sampling}
Janis Postels, Francesco Ferroni, Huseyin Coskun, Nassir Navab, and Federico
  Tombari.
\newblock Sampling-free epistemic uncertainty estimation using approximated
  variance propagation.
\newblock In \emph{Proceedings of the IEEE International Conference on Computer
  Vision}, pages 2931--2940, 2019.

\bibitem[Qiu et~al.()Qiu, Meyerson, and Miikkulainen]{qiu2019quantifying}
Xin Qiu, Elliot Meyerson, and Risto Miikkulainen.
\newblock Quantifying point-prediction uncertainty in neural networks via
  residual estimation with an {I}/{O} kernel.
\newblock \emph{ICLR 2020}.

\bibitem[Ritter et~al.(2018)Ritter, Botev, and Barber]{ritter2018scalable}
Hippolyt Ritter, Aleksandar Botev, and David Barber.
\newblock A scalable {L}aplace approximation for neural networks.
\newblock \emph{{ICLR}}, 2018.

\bibitem[Rubner et~al.(1998)Rubner, Tomasi, and Guibas]{earthmover}
Yossi Rubner, Carlo Tomasi, and Leonidas~J. Guibas.
\newblock A metric for distributions with applications to image databases.
\newblock In \emph{Proceedings of the Sixth International Conference on
  Computer Vision}, ICCV '98, page~59, USA, 1998.

\bibitem[Sicking et~al.(2019)Sicking, Kister, Fahrland, Eickeler, H{\"u}ger,
  R{\"u}ping, Schlicht, and Wirtz]{workshop_unc_realism}
Joachim Sicking, Alexander Kister, Matthias Fahrland, Stefan Eickeler, Fabian
  H{\"u}ger, Stefan R{\"u}ping, Peter Schlicht, and Tim Wirtz.
\newblock Approaching neural network uncertainty realism.
\newblock \emph{NeurIPS 2019 Workshop on Machine Learning for Autonomous
  Driving}, 2019.

\bibitem[Sicking et~al.(2020)Sicking, Akila, Wirtz, Houben, and
  Fischer]{sicking2020characteristics}
Joachim Sicking, Maram Akila, Tim Wirtz, Sebastian Houben, and Asja Fischer.
\newblock Characteristics of {M}onte {C}arlo dropout in wide neural networks.
\newblock \emph{ICML 2020 Workshop on Uncertainty and Robustness in Deep
  Learning, arXiv:2007.05434}, 2020.

\bibitem[Snoek et~al.(2019)Snoek, Ovadia, Fertig, Lakshminarayanan, Nowozin,
  Sculley, Dillon, Ren, and Nado]{snoek2019can}
Jasper Snoek, Yaniv Ovadia, Emily Fertig, Balaji Lakshminarayanan, Sebastian
  Nowozin, D~Sculley, Joshua Dillon, Jie Ren, and Zachary Nado.
\newblock Can you trust your model's uncertainty? evaluating predictive
  uncertainty under dataset shift.
\newblock In \emph{Advances in Neural Information Processing Systems}, pages
  13969--13980, 2019.

\bibitem[Srivastava et~al.(2014)Srivastava, Hinton, Krizhevsky, Sutskever, and
  Salakhutdinov]{srivastava2014dropout}
Nitish Srivastava, Geoffrey Hinton, Alex Krizhevsky, Ilya Sutskever, and Ruslan
  Salakhutdinov.
\newblock Dropout: a simple way to prevent neural networks from overfitting.
\newblock \emph{The Journal of Machine Learning Research}, 15\penalty0
  (1):\penalty0 1929--1958, 2014.

\bibitem[Stephens(1974)]{stephens1974edf}
Michael~A Stephens.
\newblock {EDF} statistics for goodness of fit and some comparisons.
\newblock \emph{Journal of the American Statistical Association}, 69\penalty0
  (347):\penalty0 730--737, 1974.

\bibitem[Villani(2008)]{villani2008optimal}
C{\'e}dric Villani.
\newblock \emph{Optimal transport: old and new}, volume 338.
\newblock Springer Science \& Business Media, 2008.

\bibitem[Walker et~al.(2016)Walker, Doersch, Gupta, and
  Hebert]{walker2016uncertain}
Jacob Walker, Carl Doersch, Abhinav Gupta, and Martial Hebert.
\newblock An uncertain future: Forecasting from static images using variational
  autoencoders.
\newblock In \emph{European Conference on Computer Vision}, pages 835--851.
  Springer, 2016.

\bibitem[Wilson et~al.(2016)Wilson, Hu, Salakhutdinov, and
  Xing]{wilson2016deep}
Andrew~Gordon Wilson, Zhiting Hu, Ruslan Salakhutdinov, and Eric~P Xing.
\newblock Deep kernel learning.
\newblock In \emph{Artificial Intelligence and Statistics}, pages 370--378,
  2016.

\end{thebibliography}

\clearpage
\appendix
{\LARGE
{\hspace{-0.75cm} Supplementary Material}
}
\vspace*{0.3cm}

\noindent
This part accompanies our paper ``\textit{A Novel Regression Loss for Non-Parametric Uncertainty Optimization}'' and provides further in-depth information.
In Section \ref{appendix:mechanics} we provide both theoretical and numerical insight into the resulting uncertainties of our loss modification.
Large parts of the empirical evaluation can be found in section \ref{appendix:empiricalStudy}, including details on the setup, data splits as well as further uncertainty measures.
As the \ours couples to the usual MSE regression loss via a hyper-parameter $\beta$ we test various values in section \ref{appendix:beta}, finding no strong correlation between result and parameter.
We close with a discussion on the relations between uncertainty measures and their respective sensitivity in section \ref{appendix:unc_measures}.

\section{Mechanics of the \ours}
\label{appendix:mechanics}

We analytically study the optimization landscape evoked by the \ours in \ref{appendix:analytical}. This analysis provides building blocks to better understand the composition of the SML-uncertainties as detailed in the remainder of this section.

\subsection{Analytical properties of the \ours}\label{appendix:analytical}
\begin{figure}[bth]
    \centering
    \includegraphics[width=0.4\textwidth]{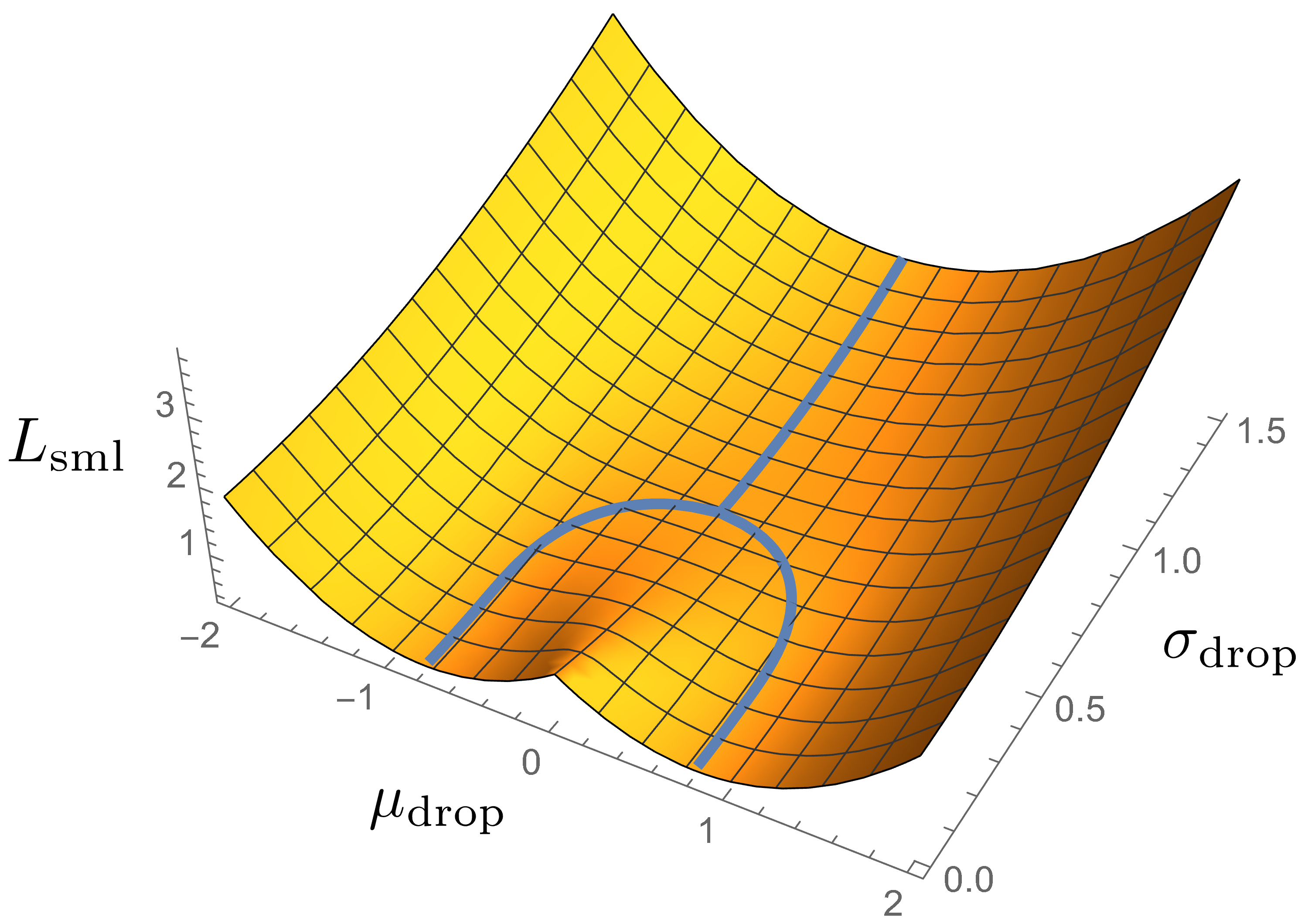}
    \caption{Shown is the value of the loss component $L_2$ as given by eq.\ (\ref{eq:l2_aleatoric_apx}) over $\mmc$ and $\smc$ describing the implicit dropout ensemble. The blue line shows the position of the minima of $L_2$ for fixed values of $\smc$. Clearly visible are the global minima at $\smc=0$ and the bifurcation at $\smc=2/\pi$.}
    \label{fig:integral_bimodular}
\end{figure}

In the following, we look closer at the behaviour of the \ours with respect to aleatoric uncertainty.
For this, we assume that the residuals, compare eq.\ (\ref{eq:mod_mc_objective_train}), are given by a Gaussian distribution with, for simplicity, $\mu_\text{Res.}=0$ and $\sigma_\text{Res.}=1$. We want to determine the resulting loss for the $L_\text{sml}$ term in eq.\ (\ref{eq:mod_mc_objective_train}) governing the uncertainty estimation of the model.
It depends on the underling distribution of the effective MC dropout distribution, which me model as $\mathcal{N}(\mmc,\smc)$ such that:
\begin{equation}
    L_\text{sml}=\int_{-\infty}^{\infty}\!\mathrm{d}y_1\mathrm{d}y_2\,{(|y_1|-|y_2|)^2\,p_1(y_1)\,p_2(y_2)}\,,
\end{equation}
where $p_1$ and $p_2$ are the Gaussian distributions discussed above.
After some calculation this yields:
\begin{align}
    L_\text{sml}=-\frac{4}{\pi} \smc \exp\left(-\frac{1}{2}\frac{\mmc^2}{\smc^2}\right) - \sqrt{\frac{8}{\pi}}\,\mmc\, \erf\left(\frac{\mmc}{\sqrt{2}\,\smc}\right) + \smc^2 + \mmc^2 + 1 \,,
    \label{eq:l2_aleatoric_apx}
\end{align}
which is visualized in Fig.\ \ref{fig:integral_bimodular}.
The two global minima can be found for $\smc=0$ and $\mmc=\pm\sqrt{2/\pi}$. However, as we model a randomized residual $y_1$ these minima do not reach zero.
We find that it is favourable to move $\mmc$ away from the network prediction of $\mu_\text{Res.}=0$, the mean of the underlying data distribution.
But, this is only the case as long as the inherent uncertainty in the dropout distribution can be brought below $\smc<2/\pi$, which is still smaller than the uncertainty of $\sigma_\text{Res.}=1$ assumed within the training data distribution.
Otherwise, it is more favourable to have $\mmc=\mu_\text{Res.}=0$.
Decomposing the uncertainty for the UCI datasets in section \ref{appendix:sigma_total_eval} showed mixed behaviour with indications for bi-modal shifts in $\mmc$ as well as improved values of $\smc$.

We already showed the effect of this bi-modality in Fig.\ \ref{fig:subnets_toydata} at the beginning of the paper, where various sub-networks where sampled. Clearly visible is a stronger variation between the networks compared to MC, but also a concentration around the two possible minima. While this Fig.\ provides a good visual estimate of $\smc$ the total uncertainty $\sigma_\text{total}$ would additionally contain the systematic shift $| \f - \langle \fsub \rangle |$. Given the roughly symmetric distribution of the sub-networks we can expect it to be comparatively small.

\subsection{Composition of the uncertainty estimate}
\label{appendix:sigma_total_eval}

\begin{figure}[bth]
    \centering
    \includegraphics[width=0.75\textwidth]{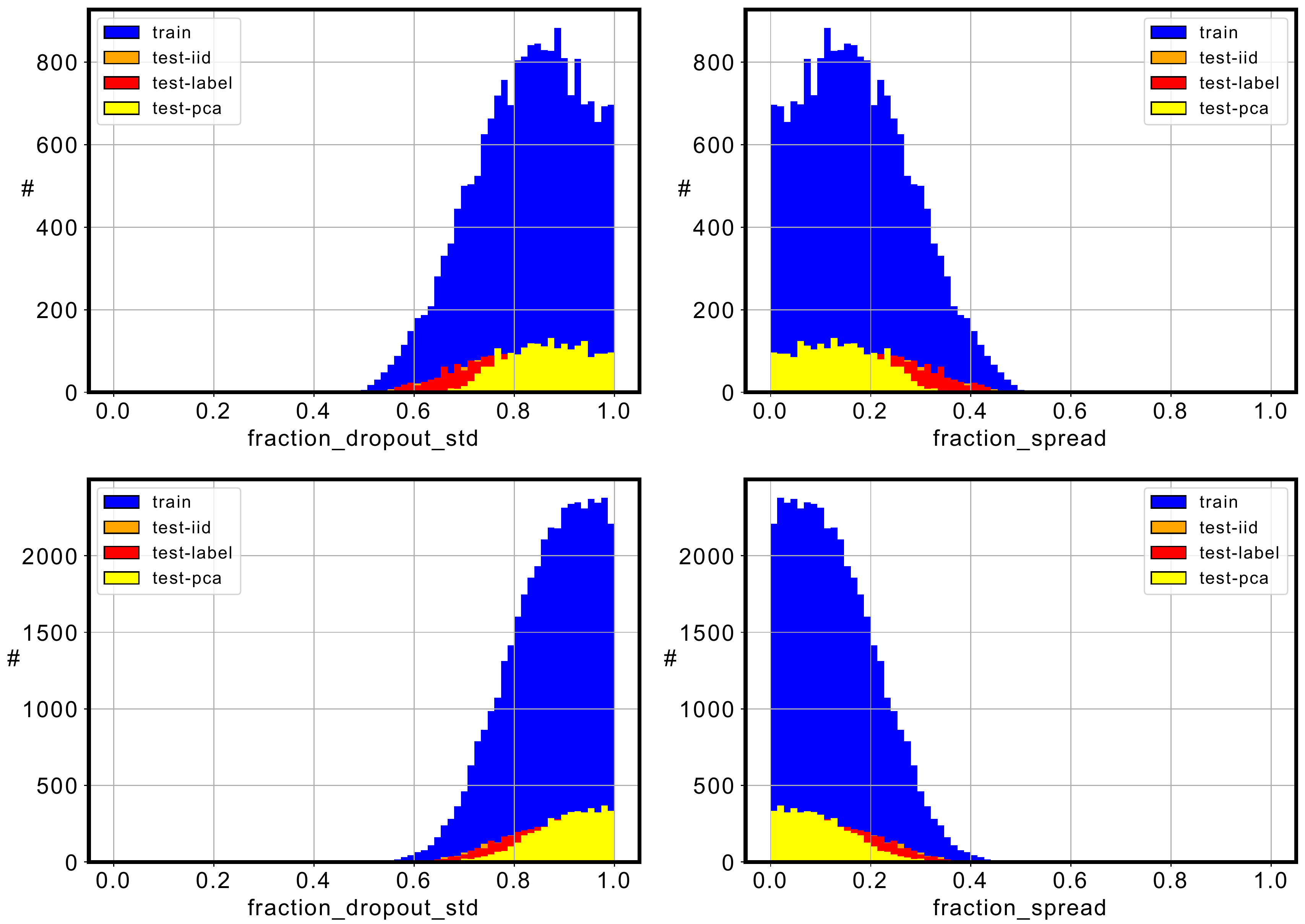}
    \caption{The \ours induces uncertainties $\sigma_\text{total}=\sigma_{\rm drop} + | f_{\btheta} - \langle f_{\bthetaTilde}\rangle |$. The relative contribution of both components ("fraction\_dropout\_std", "fraction\_spread") is shown for two exemplary datasets (top: superconduct, bottom: protein) and \iid (train: blue, test: orange) as well as non-\iid data splits (test-label: red, test-pca: yellow).}
    \label{fig:std_total_analysis}
\end{figure}

The uncertainty estimate of the \ours is comprised of two parts: $\sigma_{\rm total} = \smc + | f_{\btheta} - \langle f_{\bthetaTilde}\rangle | $.
Fig. \ref{fig:std_total_analysis} reveals that $\sigma_{\rm drop}$ contributes to more than $80\%$ of $\sigma_{\rm total}$ for the three presented datasets and for all applied data splits. A highly similar behavior can be observed for all other datasets. The analytical consideration in appendix \ref{appendix:analytical} suggests that for cases without aleatoric uncertainty the SML provides no incentive for $| f_{\btheta} - \langle f_{\bthetaTilde}\rangle | > 0$. The same holds true in the presence of aleatoric uncertainty as long as $\sigma_{\rm drop}$ is comparably large. For aleatoric uncertainty and small $\sigma_{\rm drop}$ larger $| f_{\btheta} - \langle f_{\bthetaTilde}\rangle |$ are favorable. However, as our loss is radial symmetric, all directions are equivalent and initialization and randomness determine the direction of the spread $| f_{\btheta} - \langle f_{\bthetaTilde}\rangle |$ for each individual sub-network. This symmetry leads again to a small averaged $| f_{\btheta} - \langle f_{\bthetaTilde}\rangle |$. $\sigma_{\rm drop}$ on the contrary describes the width of a bi-modal set of sub-networks in these cases.

\subsection{Detailed analysis of the two loss components}\label{appendix:two_loss_comps}

\begin{figure}[bth]
    \centering
    \includegraphics[width=1.0\textwidth]{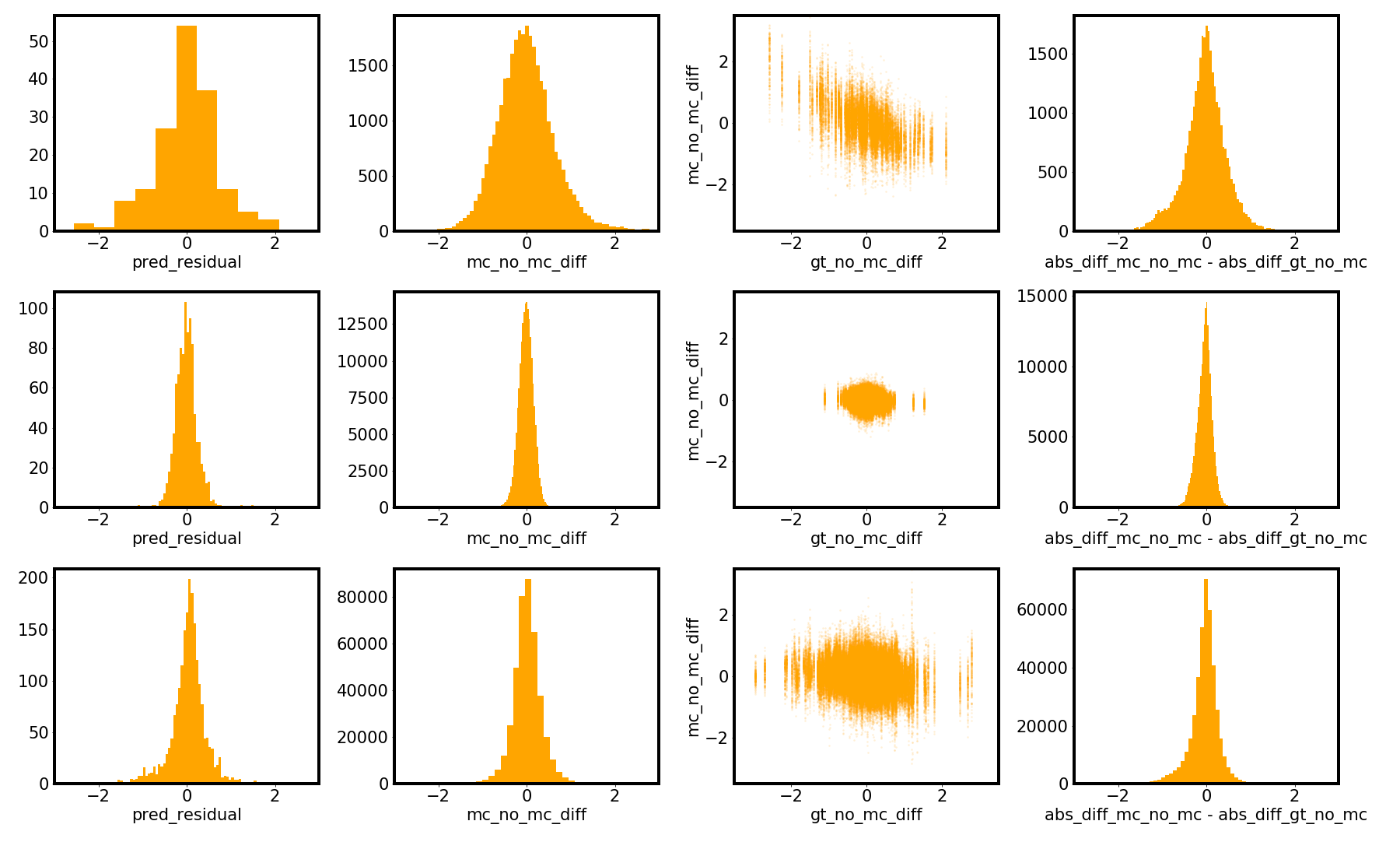}
    \caption{Visualisation of the components (columns) of the \ours for selected test datasets (rows). The prediction residual $f_{\boldsymbol{\theta}}(x_i) - y_i$  (first column), model spread $f_{\bthetaTilde}(x_i) - f_{\btheta}(x_i)$ (second column), a scatter plot of both quantities (third column) and $|f_{\bthetaTilde}(x_i) - f_{\btheta}(x_i)| - |f_{\boldsymbol{\theta}}(x_i) - y_i|$ (fourth column) are shown. The chosen datasets from top to bottom are: wine-red, power and california.}
    \label{fig:two_loss_comps}
\end{figure}

A deeper look into the structure of the \ours is possible if we investigate its behaviour component-wise. To clarify the results presented in Fig. \ref{fig:two_loss_comps}, we recall the loss structure as
\begin{equation}
    L = L_1 + L_2 = \sum_{i=1}^M \left[a_i^2 + \beta\,(|b_i| - |a_i|)^2\,\right]
\end{equation}
with $a_i = f_{\boldsymbol{\theta}}(x_i) - y_i$ and $b_i = f_{\bthetaTilde}(x_i) - f_{\btheta}(x_i)$. Histograms of the $a_i$ (Fig. \ref{fig:two_loss_comps}, first column) enable a detailed view on network performance. The uncertainty quality of the networks can be judged by studying the $L_2$ loss term more closely, namely by visualizing histograms of $|b_i|-|a_i|$ (fourth column). The second and third column zoom into $L_2$ and show histograms of the $b_i$ and scatter plots of ($b_i$,$a_i$), respectively. Only test datasets are visualized and as we applied $90\!:\!10$ train-test splits, this explains the low resolution of some histograms in the first column. All quantities involving $b_i$ require the sampling of sub-networks. We draw $200$ sub-networks. This sampling procedure ex\-plains the higher plot resolutions in columns two to four.

Qualitatively, we observe that both the $a_i$'s and $b_i$'s are centered around zero which hints at successful optimization of regression performance and of uncertainty quality. Details on how the optimization is technically realistic, can be gained from the scatter plots. They show two qualitative shapes: a `line' (first row) and a `blob' (second and third row).
For an in-detail discussion of the uni- and bi-modality of the \ours landscape see \ref{appendix:analytical}.  A `line' shape reflects that all sub-networks occupy the same minimum given a bi-modal case. Following appendix \ref{appendix:analytical}, a `blob' indicates a uni-modal case that might be evoked by large standard deviations $\sigma_{\rm drop}$.
\section{Extension to the empirical study}
\label{appendix:empiricalStudy}

Accompanying to the evaluation sketched in the body of the paper, section \ref{sec:experiments}, we provide more details on the setup, used benchmarks and measures in the following sub-section.
Further information on the experiments are given in section \ref{appendix:uci_eval}, which we extend by the measures skipped in the main text, and include a description on the used label splits.
We close with a look at the predicted uncertainties (per method) via scatter plots in section \ref{appendix:res_error}.

\subsection{Experimental setup}
\label{appendix:experimentalSetup}

The experimental setup used for the experiments is presented in three parts: the benchmark approaches we compare with, the evaluation measures we apply to quantify uncertainty, and a description of the neural networks and training procedures we employ. 

\paragraph{Benchmark approaches} We compare dropout networks trained with the SML to ar\-che\-types of uncertainty modelling, namely approximate Bayesian techniques, parametric uncertainty, and ensembling approaches.
From the first group, we pick MC dropout (abbreviated as \textbf{MC}) and its variant last-layer MC dropout (\textbf{MC-LL}).
While these dropout approaches integrate uncertainty estimation into the very structure of the network, \textit{parametric} approaches model the variance directly as the output of the neural network \citep{nix1994estimating}. Such networks typically output mean and variance of a Gaussian distribution $(\mu, \sigma)$ 
and are trained by likelihood maximization.
This approach is denoted as \textbf{PU} for parametric uncertainty. Ensembles of PU-networks \citep{lakshminarayanan2017simple}, referred to as deep ensembles, pose a widely used state-of-the-art method for uncertainty estimation \citep{snoek2019can}. Moreover, we consider ensembles of non-parametric standard networks. We refer to the latter ones as \textbf{DEs} while we call those using PU \textbf{PU-DEs}. All considered types of networks provide estimates $(\mu_i,\sigma_i)$ where $\sigma_i$ is obtained either analytically (PU), by sampling (MC, MC-LL, SML) or as an ensemble aggregate (DE, PU-DE).

\paragraph{Evaluation measures} In all experiments we evaluate both regression performance and uncertainty quality. Regression performance is quantified by the root-mean-square error (\textbf{RMSE}), $\sqrt{(1/N\,\sum_i (\mu_i-y_i)^2 }$ \citep{bishop2006pattern}. Another established metric in the uncertainty community is the (Gaussian) negative log-likelihood (\textbf{NLL}), $1/N \sum_i \big( \log\sigma_i\allowbreak + (\mu_i - y_i)^2 \allowbreak /(2 \sigma_i^2) + c \big)$, a hybrid between performance and uncertainty measure \citep{gneiting2007strictly}, see appendix \ref{appendix:nll} for a discussion.\footnote{Throughout the paper, we ignore the constant $c=\log\sqrt{2\pi}$ of the NLL.} The expected calibration error (\textbf{ECE}, \citet{Kuleshov2018})
in contrast is not biased towards well-performing models and in that sense is a pure uncertainty measure. It reads ECE $= \sum_{j=1}^B |\tilde{p}_j - 1/B|$ for B equally spaced bins in quantile space and $\tilde{p}_j = |\{r_i | q_j \leq \tilde{q}(r_i) < q_{j+1}\}|/N$ the empirical frequency of data points falling into such a bin. The normalized prediction residuals $r_i$ are defined as $r_i = (\mu_i - y_i)/\sigma_i$. Additionally, we propose to consider the \textit{Wasserstein distance of normalized prediction residuals} (\textbf{WS}). The Wasserstein distance \citep{villani2008optimal}, also known as earth mover's distance \citep{earthmover}, is a transport-based measure denoted by ($d_{\rm WS}$) between two probability densities, with Wasserstein GANs \citep{arjovsky2017wasserstein} as its most prominent application in ML. 
For ideally calibrated uncertainties, we expect $y_i \sim \mathcal{N}(\mu_i,\sigma_i)$
and therefore $r_i \sim \mathcal{N}(0,1)$. Thus we use $d_{\rm WS}(\{r_i\}_i,\mathcal{N}(0,1))$
to measure deviations from this ideal behavior. As ECE, this is a pure uncertainty measure. However, it does not use binning and can therefore resolves deviations on all scales. For example, two strongly ill-calibrated uncertainties ($r_1,r_2\gg1$, $r_1<r_2$) would result in (almost) identical ECE values while WS would resolve this difference in magnitude.

\paragraph{Technical details} All investigated neural networks have the same architecture, $2$ hidden layers of width $50$, and ReLu activations \citep{relu3}. For all dropout-based methods (MC, MC-LL, SML) we set the drop rate to $p = 0.1$. Like MC, SML-trained networks apply Bernoulli dropout to all hidden activations. In the case of MC-LL the dropout is only applied to the last hidden layer. For ensemble methods (DE, DE-PU) we employ $5$ networks.
For PE networks, we normalize the $\sigma$ value using softplus \citep{relu3} and optimzie the NLL instead of the MSE. 
For the optimization  of all NNs we use the ADAM-optimizer \citep{kingma2014adam} with a learning rate of $0.001$.
For `california', the learning rate is reduced to $0.0001$ as training of PU and PU-DE is unstable using the standard setup.
Additionally, we apply standard normalization to the input and output features of all datasets to enable better comparability.

Number of epochs trained and amount of cross validation differs by the training-set size.
We categorize the datasets as follows:  small datasets \{yacht, diabetes, boston, energy, concrete, wine-red\,\}, large datasets \{abalone, power, naval, california, superconduct, protein\,\} and very large datasets \{year\,\}. For small datasets, NNs are trained for $1,000$ epochs using mini-batches of size $100$. All results are 10-fold cross validated. For large datasets, we train for $150$ epochs and apply 5-fold cross validation. We keep this large-dataset setting for the very large `year' dataset but increase mini-batch size to $500$.

All experiments are conducted on \texttt{Core Intel(R) Xeon(R) Gold 6126} CPUs.
Conducting the described experiments with cross validation on one CPU takes $80\,h$.

For SML it turns out that as long as $0\!<\!\beta\!<\!1$, the actual value of $\beta$ has only a limited influence on the optimization result, see appendix \ref{appendix:beta} for details. Larger $\beta$-values can however favour uncertainty optimization at an expanse of task performance. Throughout the body of the paper we use a conservative value of $\beta=0.5$.

\subsection{RMSEs, NLLs and systematic evaluation}\label{appendix:uci_eval}

This sub-section provides further details on our experiments covering: an overview on the datasets and splits used for the data-shift studies, further uncertainty measure evaluations (RMSE, NLL, WS),  and close with a discussion of the weaker SML results.

\begin{figure}[b!th]
    \centering 
    \includegraphics[trim=50 180 50 60, clip, width=0.85\textwidth]{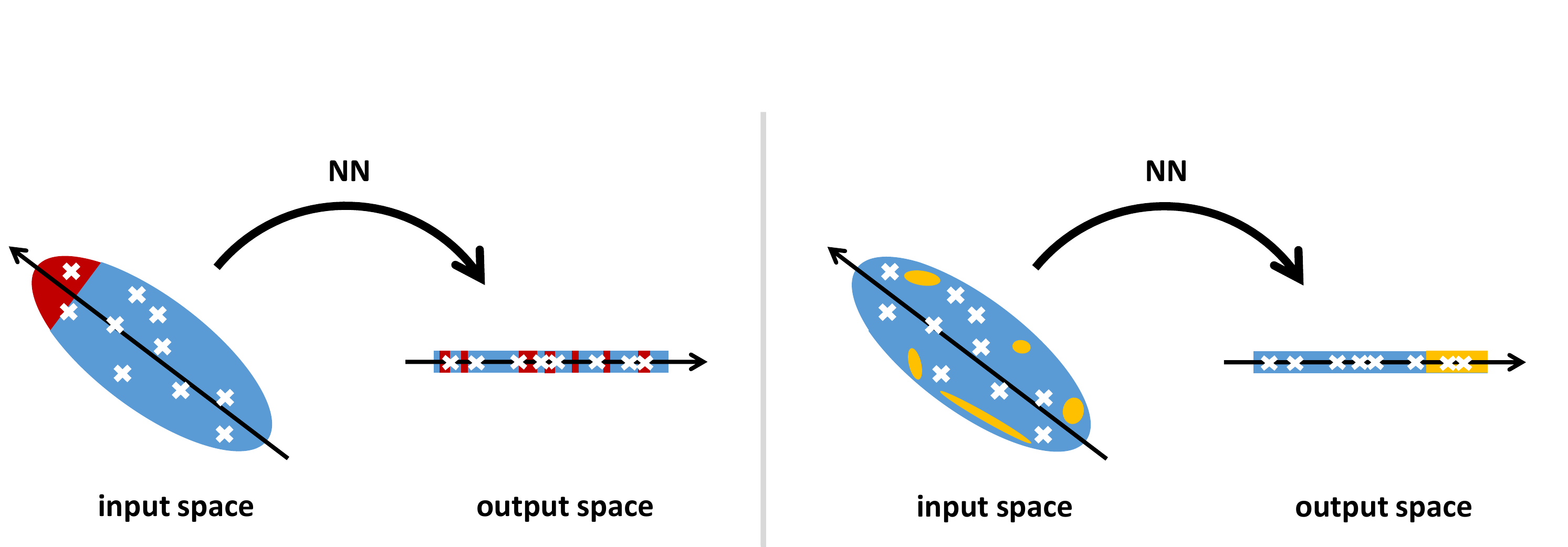}
    \caption{Scheme of two non-\iid splits: a PCA-based split in input space (left) and label-based split in output space (right). While datasets appear to be convex here, they are (most likely) not in reality.}
    \label{fig:pca_label_split}
\end{figure}

\paragraph{Datasets and data splits}
For the regression data, Table \ref{tab:uci_data_preproc} provides details on dataset references, preprocessing and basic statistics.
Extrapolation and interpolation data-shifts are, technically, introduced by applying non-\iid~(independent and identically distributed) data splits. Natural candidates for such non-\iid splits are splits along the main directions of data in input and output space, respectively. Here, we consider 1D regression tasks. Therefore, output-based splits are simply done on a scalar label variable (see Fig. \ref{fig:pca_label_split}, right). We call such a split \textit{label-based} (for a comparable split, see, \egnows, \citet{foong2019between}). In input space, the first component of a principal component analysis (PCA) provides a natural direction (see Fig. \ref{fig:pca_label_split}, left). The actual \textit{PCA-split} is then based on projections of the data points onto this first PCA-component.\footnote{Note that these projections are only considered for data splitting, they are not used for model training.} Splitting data along such an direction in input or output space in \eg $10$ equally large chunks, creates $2$ \textit{outer} data chunks and $8$ \textit{inner} data chunks. Training a model on $9$ of these chunks such that the remaining chunk for evaluation is an inner chunk is called data \textit{interpolation}. If the remaining test chunk is an outer chunk, it is data \textit{extrapolation}. We introduce this distinction as extrapolation is expected to be considerably more difficult than `bridging' between feature combinations that were seen during training.
\begin{table}[bht] 
\centering
\caption{Details on UCI regression datasets. Ground truth (gt) is partially pre-processed to match the 1D regression setup.}\label{tab:uci_data_preproc}
\vspace{0.8em}
\begin{tabular}{l r r l l}
\toprule
  {dataset} & {\# features} & {\# datapoints} & {reference} & {remarks} \\
\bottomrule
  \addlinespace[0.5em]
yacht       &  6    &  308      & UCI  & \\
diabetes    &  7    &  442      & sklearn  & \\
boston      &  13   &  506      & sklearn  & \\
energy      &  8    &  768      & UCI  & only "cooling load" gt used\\
concrete    &  8    &  1030     & UCI  & \\
wine-red    &  11   &  1599     & UCI  & \\
abalone    &  7    &  4176     & UCI  & $1^{\rm st}$ feature is ignored\\
power       &  4    &  9568     & UCI  & \\
naval       &  16   &  11934    & UCI  & using only "turbine" gt\\
california  &  8    &  20640    & sklearn  & \\
superconduct&  81   &  21263    & UCI  & \\
protein     &  9    &  45730    & UCI  & \\
year        &  90   &  515345   & UCI  & \\
\bottomrule
  \addlinespace[0.8em]
\end{tabular}
\end{table}

\begin{figure}[ptbh]
    \centering
    \includegraphics[width=1.0\textwidth]{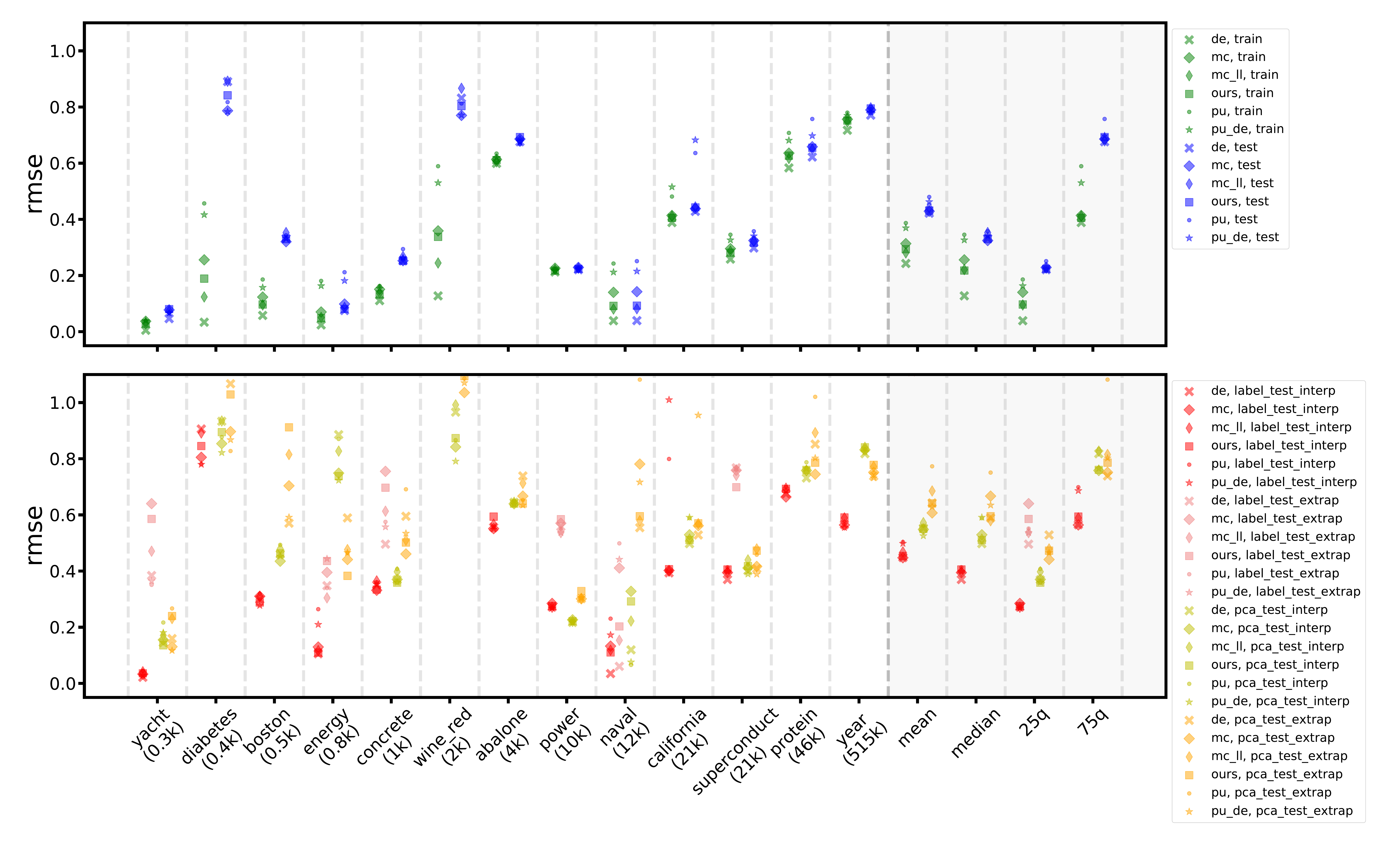}
    \vspace{-3em}
    \caption{Root-mean-square errors (RMSEs) for 13 UCI regression datasets under \iid conditions (top) and under data shift (bottom). Uncertainty methods are encoded via plot marker, data splits via color. Each plot point corresponds to a cross-validated trained network. Summarizing statistics (rhs) are indicated by a light grey background.}
    \label{fig:uci_rmse}
\end{figure}

\paragraph{Regression quality}
First, we consider regression performance (see top panel of Fig. \ref{fig:uci_rmse}).
Averaging the RMSE values over the considered $13$ datasets (`mean' column) yields almost identical results for all uncertainty methods. The only exceptions pose PU and PU-DE with larger train data RMSEs which could be due to NLL optimization favoring to adapt variance rather than mean. However, this regularizing NLL-training comes along with a smaller generalization gap, leading to competitive test RMSEs. Next, we investigate model performance under data shift, 
visualized in the bottom panel of Fig. \ref{fig:uci_rmse}. Again, regression quality is comparable between all methods. As expected, performances under data shift are worse compared to those on \iid test sets.
\begin{figure}[pbth]
    \includegraphics[width=1.0\textwidth]{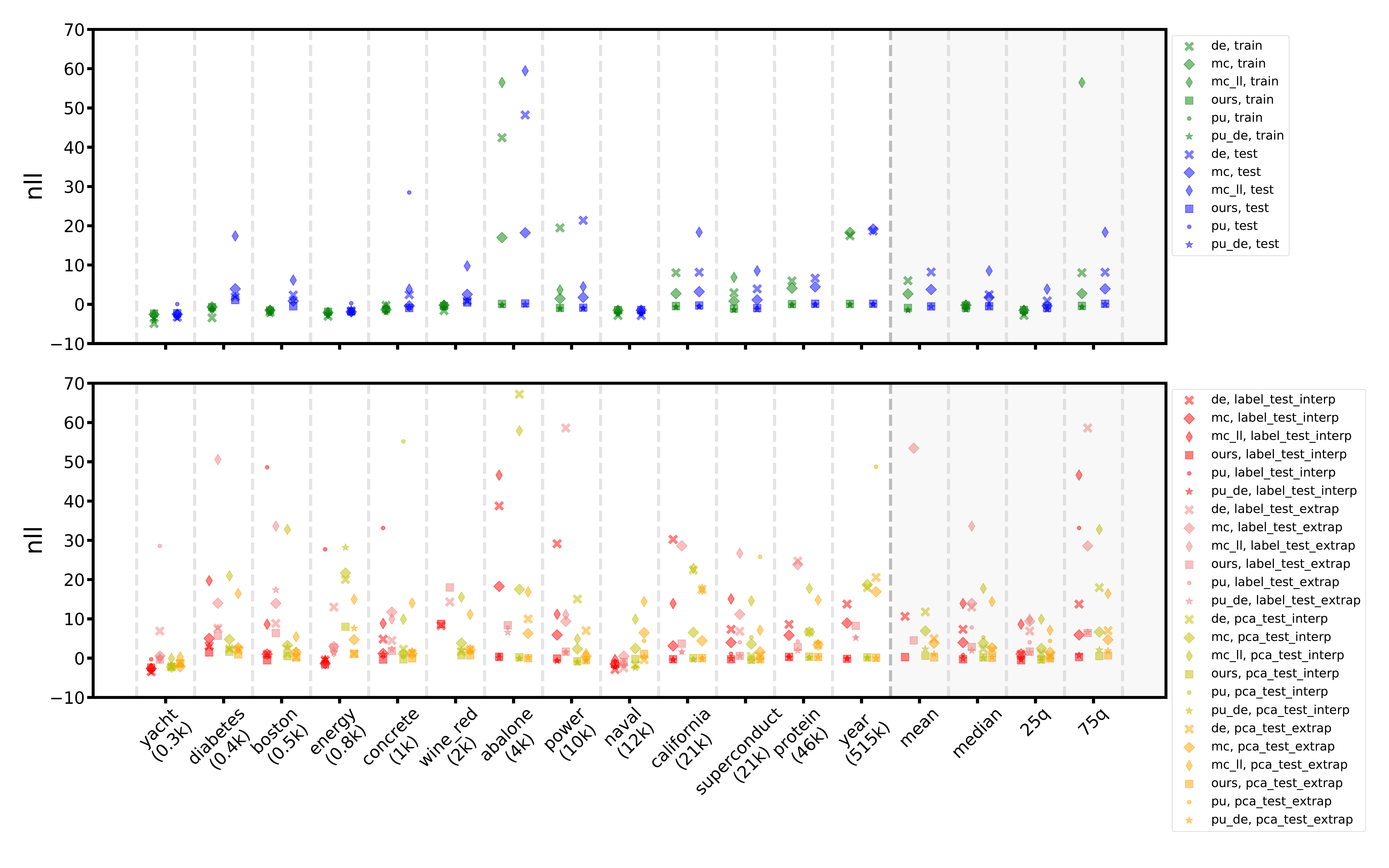}
    \vspace{-3em}
    \caption{Negative log-likelihoods (NLLs) for 13 UCI regression datasets under \iid conditions (top) and under data shift (bottom). Uncertainty methods are encoded via plot marker, data splits via color. Each plot point corresponds to a cross-validated trained network. Summarizing statistics (rhs) are indicated by a light grey background.}
    \label{fig:uci_nll}
\end{figure}

\paragraph{Negative log-likelihoods}
For NLL, results are less balanced compared to RMSE (see Fig. \ref{fig:uci_nll}). PU-DE and the SML-trained network reach comparably small average values, followed by MC and DE. The average NLL values of MC-LL and PU are above the upper plot limit indicating a rather weak stability of these methods. On PCA-interpolate and PCA-extrapolate test sets, again PU-DE and SML-trained networks perform best. On label-interpolate and label-extrapolate test sets, SML-trained networks take the first place with a large margin. The mean NLL values of most other approaches are above the upper plot limit. Note that median results (the column next to `mean') are not as widely spread and PU-DE and SML perform comparably well. These qualitative differences between mean and median behavior indicate that most methods perform poorly `once in a while'. A noteworthy observation as \textit{stability across a variety of data shifts and datasets} can be seen as a crucial requirement for an uncertainty method. SML-based models yield the highest stability in that sense \wrt NLL.

\begin{figure}[bht]
    \centering
    \includegraphics[width=1.0\textwidth]{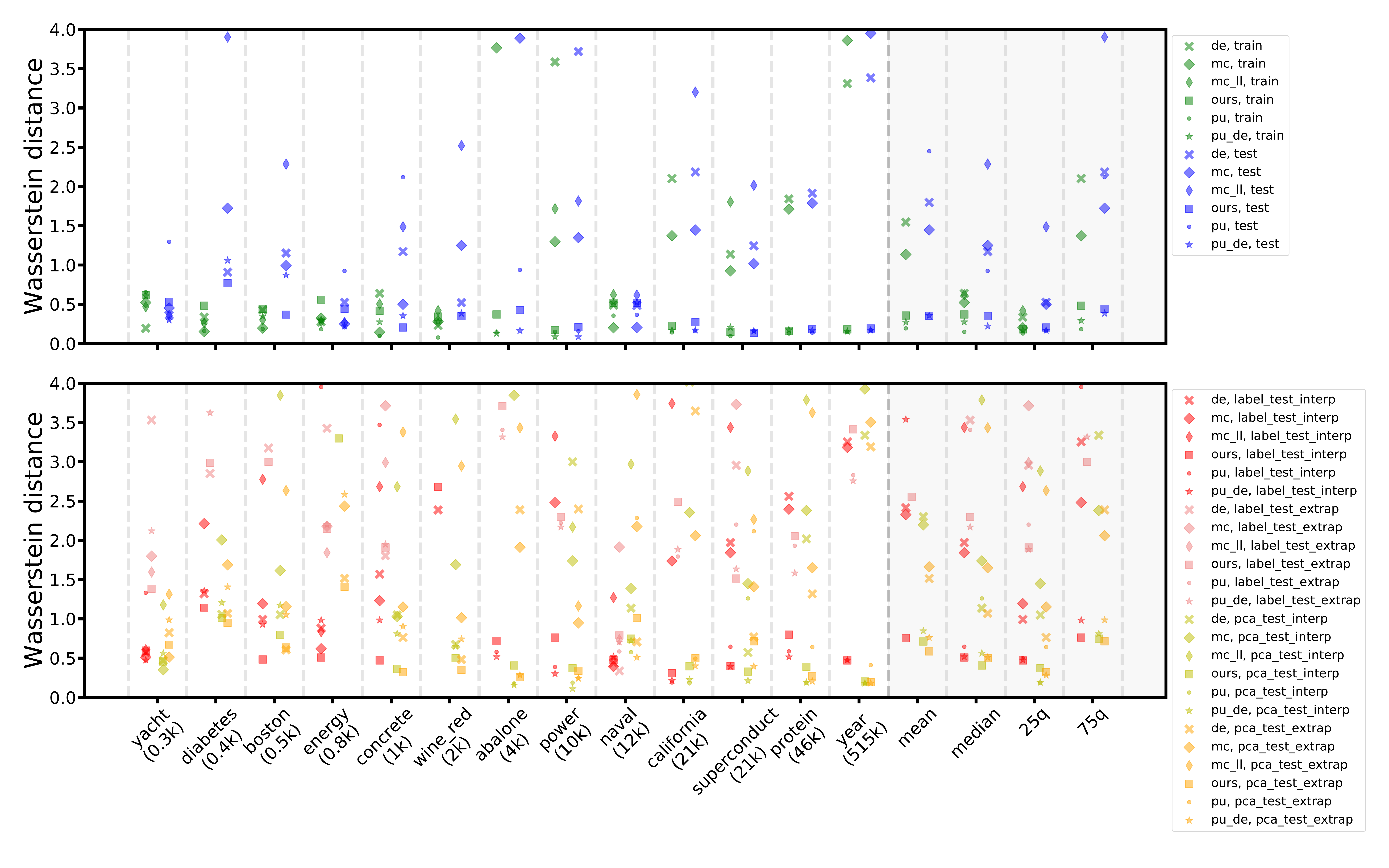}
    \vspace{-3em}
    \caption{Wasserstein distances for 13 UCI regression datasets under \iid conditions (top) and under data shift (bottom). Uncertainty methods are encoded via plot marker, data splits via color. Summarizing statistics (rhs) are indicated by a light grey background.}
    \label{fig:uci_ws}
\end{figure}

\paragraph{Wasserstein distances}
Studying Wasserstein distances, we again observe equally strong results for PU-DE and SML on train and test data (see column `mean' in top panel of Fig.\ \ref{fig:uci_ws}). PU in contrast possesses a large generalization gap thus yielding weak test set performances. MC, MC-LL, and DE behave consistently weak on train and test sets with MC-LL even falling out of plot range. Under data shift (bottom panel of Fig. \ref{fig:uci_ws}), the picture remains similar. PU-DE and SML are in the lead and comparably strong with the exception of PU-DE on label-interpolate and label-extrapolate test data (`mean' column). As for NLL, we find these mean values of PU-DE to be significantly above the respective median values indicating again weaknesses in the stability of parametric ensembles.

\paragraph{Slight overestimation of small uncertainties for SML} The \ours yields weak results on `yacht', `energy' and `naval', the three easiest datasets if measured by test set RMSE, compare Fig.\ \ref{fig:uci_nll}.
On these datasets neither aleatoric uncertainty nor modelling residuals play a mayor role. In such cases, the \ours seems to cause slightly overshooting uncertainty estimates (compare edges of Fig. \ref{fig:subnets_toydata} for a visual clue), likely due to its sub-network `repulsion'.
Back-propagating not only through $\fsub$ but also through the full network $\f$ in L$_{sml}$ might mitigate this effect. In practice, slight overestimations of small uncertainties might be acceptable. In contrast, our method performs consistently strong on all more challenging datasets (`california', `superconduct', `protein', `year'). A beneficial characteristic for virtually any real-world task.

\begin{table}

\centering
\caption{Regression performance and uncertainty quality of networks with different uncertainty mechanisms. The scores are calculated on the test sets of 13 UCI datasets.}\label{tab:apx_eval_uci}
\vspace{0.8em}
\begin{adjustbox}{max width=0.65\textwidth}
\begin{tabular}{l l r r r r r r}
\toprule
\textbf{measure} & \textbf{dataset} & \textbf{MC} & \textbf{MC-LL} & \textbf{Ours} & \textbf{PU} & \textbf{PU-DE} & \textbf{DE} \\
\midrule

RMSE ($\downarrow$) &   yacht &   $ 0.08 $ &   $ 0.07 $ &  $ 0.08 $ &  $ 0.07 $ &   $ 0.07 $ &   $ 0.05 $ \\
NLL ($\downarrow$)  &   yacht &  $ -2.53 $ &  $ -2.68 $ &  $ -2.3 $ &  $ 0.05 $ &  $ -3.53 $ &  $ -3.26 $ \\
ECE ($\downarrow$)  &   yacht &   $ 0.96 $ &   $ 0.78 $ &  $ 1.10 $ &  $ 0.90 $ &   $ 0.74 $ &   $ 0.63 $ \\
WS ($\downarrow$)   &   yacht &   $ 0.45 $ &   $ 0.36 $ &  $ 0.53 $ &  $ 1.30 $ &   $ 0.30 $ &   $ 0.36 $ \\
\midrule
\addlinespace[0.4em]
  
RMSE ($\downarrow$) &  diabetes &  $ 0.79 $ &  $ 0.89 $ &  $ 0.84 $ &    $ 0.82 $ &  $ 0.78 $ &  $ 0.89 $ \\
NLL ($\downarrow$)  &  diabetes &  $ 3.93 $ &  $ 17.4 $ &  $ 1.10 $ &  $ 316.53 $ &  $ 2.14 $ &  $ 1.90 $ \\
ECE ($\downarrow$)  &  diabetes &  $ 0.99 $ &  $ 1.26 $ &  $ 0.72 $ &    $ 1.15 $ &  $ 0.78 $ &  $ 0.71 $ \\
WS ($\downarrow$)   &  diabetes &  $ 1.72 $ &  $ 3.90 $ &  $ 0.77 $ &    $ 9.24 $ &  $ 1.06 $ &  $ 0.91 $ \\
\midrule
\addlinespace[0.4em]

RMSE ($\downarrow$) &  boston &  $ 0.32 $ &  $ 0.35 $ &   $ 0.33 $ &   $ 0.34 $ &  $ 0.33 $ &  $ 0.33 $ \\
NLL ($\downarrow$)  &  boston &  $ 0.85 $ &  $ 6.15 $ &  $ -0.48 $ &  $ 144.2 $ &  $ 1.04 $ &  $ 2.35 $ \\
ECE ($\downarrow$)  &  boston &  $ 0.76 $ &  $ 1.05 $ &   $ 0.53 $ &   $ 1.13 $ &  $ 0.70 $ &  $ 0.69 $ \\
WS ($\downarrow$)   &  boston &  $ 0.99 $ &  $ 2.29 $ &   $ 0.37 $ &   $ 5.57 $ &  $ 0.87 $ &  $ 1.15 $ \\
\midrule
\addlinespace[0.4em]

RMSE ($\downarrow$) &  energy &   $ 0.10 $ &  $ 0.09 $ &   $ 0.08 $ &  $ 0.21 $ &   $ 0.18 $ &   $ 0.08 $ \\
NLL ($\downarrow$)  &  energy &  $ -1.93 $ &  $ -1.7 $ &  $ -1.83 $ &  $ 0.26 $ &  $ -2.09 $ &  $ -1.65 $ \\
ECE ($\downarrow$)  &  energy &   $ 0.54 $ &  $ 0.45 $ &   $ 0.83 $ &  $ 0.68 $ &   $ 0.42 $ &   $ 0.52 $ \\
WS ($\downarrow$)   &  energy &   $ 0.25 $ &  $ 0.27 $ &   $ 0.44 $ &  $ 0.93 $ &   $ 0.22 $ &   $ 0.52 $ \\
\midrule
\addlinespace[0.4em]

RMSE ($\downarrow$) &  concrete &  $ 0.25 $ &  $ 0.27 $ &   $ 0.25 $ &   $ 0.29 $ &   $ 0.26 $ &  $ 0.25 $ \\
NLL ($\downarrow$)  &  concrete &  $ -0.4 $ &  $ 3.86 $ &  $ -0.93 $ &  $ 28.47 $ &  $ -0.72 $ &  $ 2.45 $ \\
ECE ($\downarrow$)  &  concrete &  $ 0.48 $ &  $ 0.73 $ &   $ 0.44 $ &   $ 0.75 $ &   $ 0.41 $ &  $ 0.65 $ \\
WS ($\downarrow$)   &  concrete &  $ 0.50 $ &  $ 1.49 $ &   $ 0.20 $ &   $ 2.12 $ &   $ 0.35 $ &  $ 1.17 $ \\
\midrule
\addlinespace[0.4em]

RMSE ($\downarrow$) &  wine-red &  $ 0.77 $ &  $ 0.87 $ &  $ 0.80 $ &      $ 0.81 $ &  $ 0.77 $ &  $ 0.83 $ \\
NLL ($\downarrow$)  &  wine-red &  $ 2.53 $ &  $ 9.76 $ &  $ 0.49 $ &  $ 14572.96 $ &  $ 0.94 $ &  $ 0.87 $ \\
ECE ($\downarrow$)  &  wine-red &  $ 0.73 $ &  $ 0.93 $ &  $ 0.41 $ &      $ 0.61 $ &  $ 0.37 $ &  $ 0.41 $ \\
WS ($\downarrow$)   &  wine-red &  $ 1.25 $ &  $ 2.52 $ &  $ 0.35 $ &     $ 10.59 $ &  $ 0.38 $ &  $ 0.52 $ \\
\midrule
\addlinespace[0.4em]

RMSE ($\downarrow$) &  abalone &   $ 0.69 $ &   $ 0.68 $ &  $ 0.69 $ &    $ 0.67 $ &   $ 0.68 $ &   $ 0.68 $ \\
NLL ($\downarrow$)  &  abalone &  $ 18.21 $ &  $ 59.45 $ &  $ 0.24 $ &  $ 610.84 $ &  $ -0.07 $ &  $ 48.21 $ \\
ECE ($\downarrow$)  &  abalone &   $ 1.29 $ &   $ 1.44 $ &  $ 0.38 $ &    $ 0.27 $ &   $ 0.29 $ &   $ 1.39 $ \\
WS ($\downarrow$)   &  abalone &   $ 3.89 $ &   $ 6.85 $ &  $ 0.43 $ &    $ 0.94 $ &   $ 0.16 $ &   $ 5.79 $ \\
\midrule
\addlinespace[0.4em]

RMSE ($\downarrow$) &   naval &   $ 0.14 $ &   $ 0.08 $ &   $ 0.09 $ &   $ 0.25 $ &   $ 0.22 $ &   $ 0.04 $ \\
NLL ($\downarrow$)  &   naval &  $ -1.51 $ &  $ -1.82 $ &  $ -1.45 $ &  $ -2.43 $ &  $ -2.37 $ &  $ -2.86 $ \\
ECE ($\downarrow$)  &   naval &   $ 0.37 $ &   $ 0.64 $ &   $ 0.94 $ &   $ 0.56 $ &   $ 0.98 $ &   $ 0.85 $ \\
WS ($\downarrow$)   &   naval &   $ 0.20 $ &   $ 0.62 $ &   $ 0.52 $ &   $ 0.37 $ &   $ 0.52 $ &   $ 0.48 $ \\
\midrule
\addlinespace[0.4em]

RMSE ($\downarrow$) &   power &  $ 0.23 $ &  $ 0.23 $ &   $ 0.22 $ &   $ 0.23 $ &   $ 0.22 $ &   $ 0.22 $ \\
NLL ($\downarrow$)  &   power &  $ 1.77 $ &  $ 4.47 $ &  $ -0.87 $ &  $ -0.97 $ &  $ -1.02 $ &  $ 21.37 $ \\
ECE ($\downarrow$)  &   power &  $ 0.79 $ &  $ 0.89 $ &   $ 0.18 $ &   $ 0.17 $ &   $ 0.15 $ &   $ 1.18 $ \\
WS ($\downarrow$)   &   power &  $ 1.35 $ &  $ 1.81 $ &   $ 0.21 $ &   $ 0.16 $ &   $ 0.09 $ &   $ 3.72 $ \\
\midrule
\addlinespace[0.4em]

RMSE ($\downarrow$) &  california &  $ 0.44 $ &   $ 0.44 $ &   $ 0.44 $ &   $ 0.64 $ &   $ 0.68 $ &  $ 0.43 $ \\
NLL ($\downarrow$)  &  california &  $ 3.21 $ &  $ 18.34 $ &  $ -0.28 $ &  $ -0.48 $ &  $ -0.58 $ &  $ 8.15 $ \\
ECE ($\downarrow$)  &  california &  $ 0.77 $ &   $ 1.07 $ &   $ 0.24 $ &   $ 0.24 $ &   $ 0.27 $ &  $ 0.90 $ \\
WS ($\downarrow$)   &  california &  $ 1.45 $ &   $ 3.20 $ &   $ 0.27 $ &   $ 0.17 $ &   $ 0.17 $ &  $ 2.18 $ \\
\midrule
\addlinespace[0.4em]

RMSE ($\downarrow$) &  superconduct &  $ 0.32 $ &  $ 0.32 $ &   $ 0.32 $ &   $ 0.36 $ &   $ 0.34 $ &  $ 0.30 $ \\
NLL ($\downarrow$)  &  superconduct &  $ 1.13 $ &  $ 8.51 $ &  $ -0.96 $ &  $ -0.87 $ &  $ -1.27 $ &  $ 3.93 $ \\
ECE ($\downarrow$)  &  superconduct &  $ 0.59 $ &  $ 0.74 $ &   $ 0.20 $ &   $ 0.15 $ &   $ 0.25 $ &  $ 0.55 $ \\
WS ($\downarrow$)   &  superconduct &  $ 1.02 $ &  $ 2.01 $ &   $ 0.14 $ &   $ 0.16 $ &   $ 0.16 $ &  $ 1.25 $ \\
\midrule
\addlinespace[0.4em]

RMSE ($\downarrow$) &  protein &  $ 0.66 $ &       $ 0.65 $ &  $ 0.66 $ &  $ 0.76 $ &   $ 0.70 $ &  $ 0.62 $ \\
NLL ($\downarrow$)  &  protein &  $ 4.45 $ &  $ 1.4\times10^6 $ &  $ 0.12 $ &  $ 0.02 $ &  $ -0.11 $ &  $ 6.65 $ \\
ECE ($\downarrow$)  &  protein &  $ 0.89 $ &       $ 1.07 $ &  $ 0.24 $ &  $ 0.22 $ &   $ 0.30 $ &  $ 0.84 $ \\
WS ($\downarrow$)   &  protein &  $ 1.79 $ &  $ 7.9\times10^5 $ &  $ 0.18 $ &  $ 0.14 $ &   $ 0.17 $ &  $ 1.91 $ \\
\midrule
\addlinespace[0.4em]

RMSE ($\downarrow$) &    year &   $ 0.79 $ &       $ 0.80 $ &  $ 0.79 $ &  $ 0.79 $ &   $ 0.78 $ &   $ 0.77 $ \\
NLL ($\downarrow$)  &    year &  $ 19.15 $ &  $ 5.7\times10^5 $ &  $ 0.12 $ &  $ 0.05 $ &  $ -0.01 $ &  $ 18.69 $ \\
ECE ($\downarrow$)  &    year &   $ 1.33 $ &       $ 1.48 $ &  $ 0.24 $ &  $ 0.27 $ &   $ 0.28 $ &   $ 1.19 $ \\
WS ($\downarrow$)   &    year &   $ 3.95 $ &   $ 6.7\times10^5 $ &  $ 0.19 $ &  $ 0.17 $ &   $ 0.17 $ &   $ 3.38 $ \\

\bottomrule
\end{tabular}
\end{adjustbox}

\end{table}

\subsection{Residual-uncertainty scatter plots}
\label{appendix:res_error}

\begin{figure}[htb]
    \centering
    \includegraphics[width=0.7\textwidth]{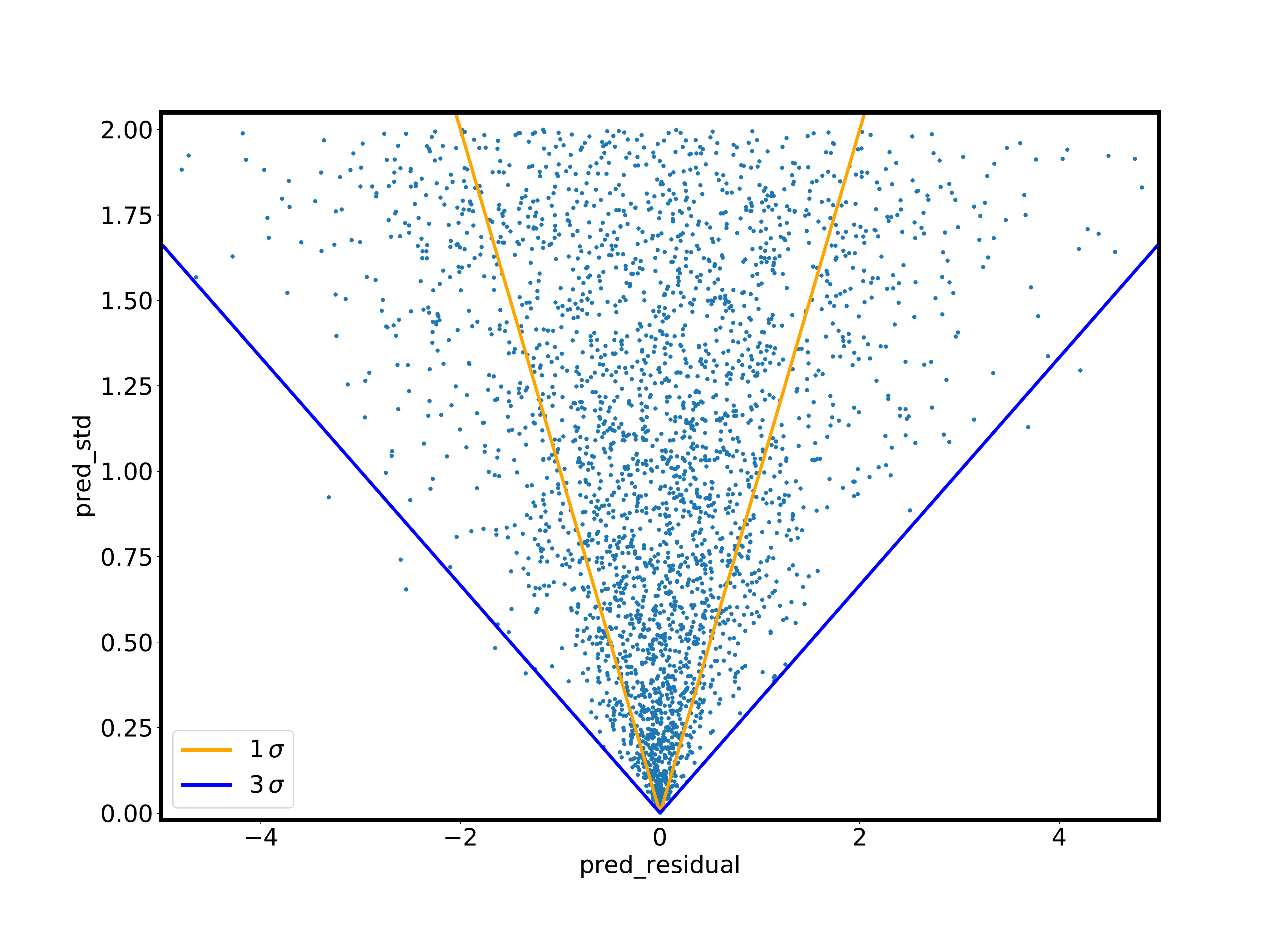}
    \caption{Prediction residuals (x-axis) and predictive uncertainty (y-axis) for a \textit{hypothetical} ideal uncertainty mechanism. The Gaussian errors are matched by Gaussian uncertainty predictions at the exact same scale. $68.3\%$ of all uncertainty estimates (plot points) lie above the orange $1\sigma$-lines and $99.7 \%$ of them above the blue $3\sigma$-lines.}
    \label{fig:residual_error_gauss}
\end{figure}

\begin{figure}[bth]
    \centering
    \includegraphics[width=1.0\textwidth]{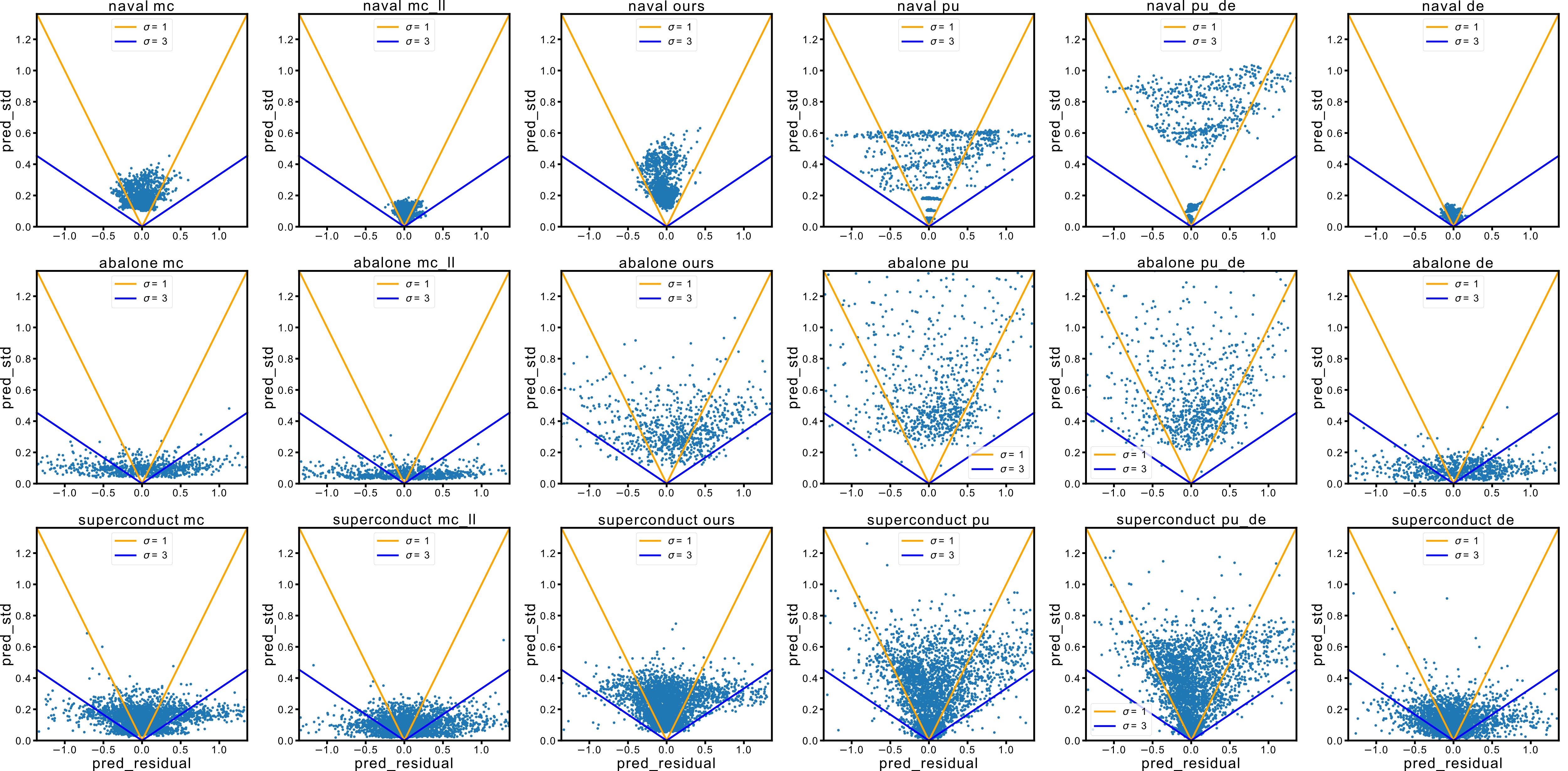}
    \caption{Prediction residuals (respective x-axis) and predictive uncertainty (respective \mbox{y-axis}) for different uncertainty mechanisms (columns) and datasets (rows). Each light blue dot in each plot corresponds to one test data point. Realistic uncertainty estimates should lie mostly above the blue $3\sigma$-lines. The datasets naval, abalone and superconduct are shown, from top to bottom.}
    \label{fig:appendix_uci_scatter}
\end{figure}

Visual inspection of uncertainties can be helpful to understand their qualitative behaviour.
We scatter model residuals $\mu_i - y_i$ (respective x-axis in Fig.\ \ref{fig:appendix_uci_scatter}) against model uncertainties $\sigma_i$ (resp. y-axis in Fig.\ \ref{fig:appendix_uci_scatter}).
For a \textit{hypothetical ideal} uncertainty mechanism, we expect $(y_i -\mu_i) \sim \mathcal{N}(0,\sigma_i)$, 
\ie model residuals following the predictive uncertainty distribution.
More concretely, $68.3\%$ of all $(y_i -\mu_i)$ would lie within the respective interval $[-\sigma_i,\sigma_i]$ and 99.7\% of all $(y_i -\mu_i)$ within $[-3\,\sigma_i, 3\,\sigma_i]$. Fig.\ \ref{fig:residual_error_gauss} visualizes this hypothetical ideal.
Geometrically, the described Gaussian properties imply that $99.7\%$ of all scatter points, \mbox{\eg in} Fig.\ \ref{fig:appendix_uci_scatter} should lie above the blue $3\sigma$ lines and $68.3\%$ of them above the yellow $1\sigma$ lines. For `abalone' test data (third row of Fig. \ref{fig:appendix_uci_scatter}), PU and SML qualitatively fulfil this requirement while MC and DE tend to underestimate uncertainties. This finding is in accordance with our systematic evaluation.
For abalone and superconduct, we qualitatively find PU, PU-DE and SML-trained networks to provide more realistic uncertainties compared to MC, MC-LL and DE (see Fig. \ref{fig:appendix_uci_scatter}). The naval dataset poses an exception in this regard as all uncertainty methods lead to comparably convincing uncertainty estimates. The small test RMSEs of all methods on naval (see Fig. \ref{fig:uci_rmse}) indicate relatively small aleatoric uncertainties and model residuals. Epistemic uncertainty might thus be a key driving factor and coherently MC, MC-LL and DE perform well.

The hypothetical ideal residual-uncertainty scatter plot we use in Fig.\ \ref{fig:residual_error_gauss} is generated as follows: We draw $3000$ standard deviations $\sigma_i \sim \mathcal{U}(0,2)$ and sample residuals $r_i$ from the respective normal distributions, $r_i \sim \mathcal{N}(0,\sigma_i)$. The pairs $(r_i,\sigma_i)$ are visualized. By construction, uncertainty estimates now ideally match residuals in a distributional sense. But even in this perfect case, Pearson correlation between uncertainty estimates and absolute residuals is only approximately $55\%$.

\section{Stability \wrt hyper-parameter \texorpdfstring{$\beta$}{beta}}
\label{appendix:beta}

\begin{figure}[ptbh]
    \centering
    \includegraphics[width=\textwidth]{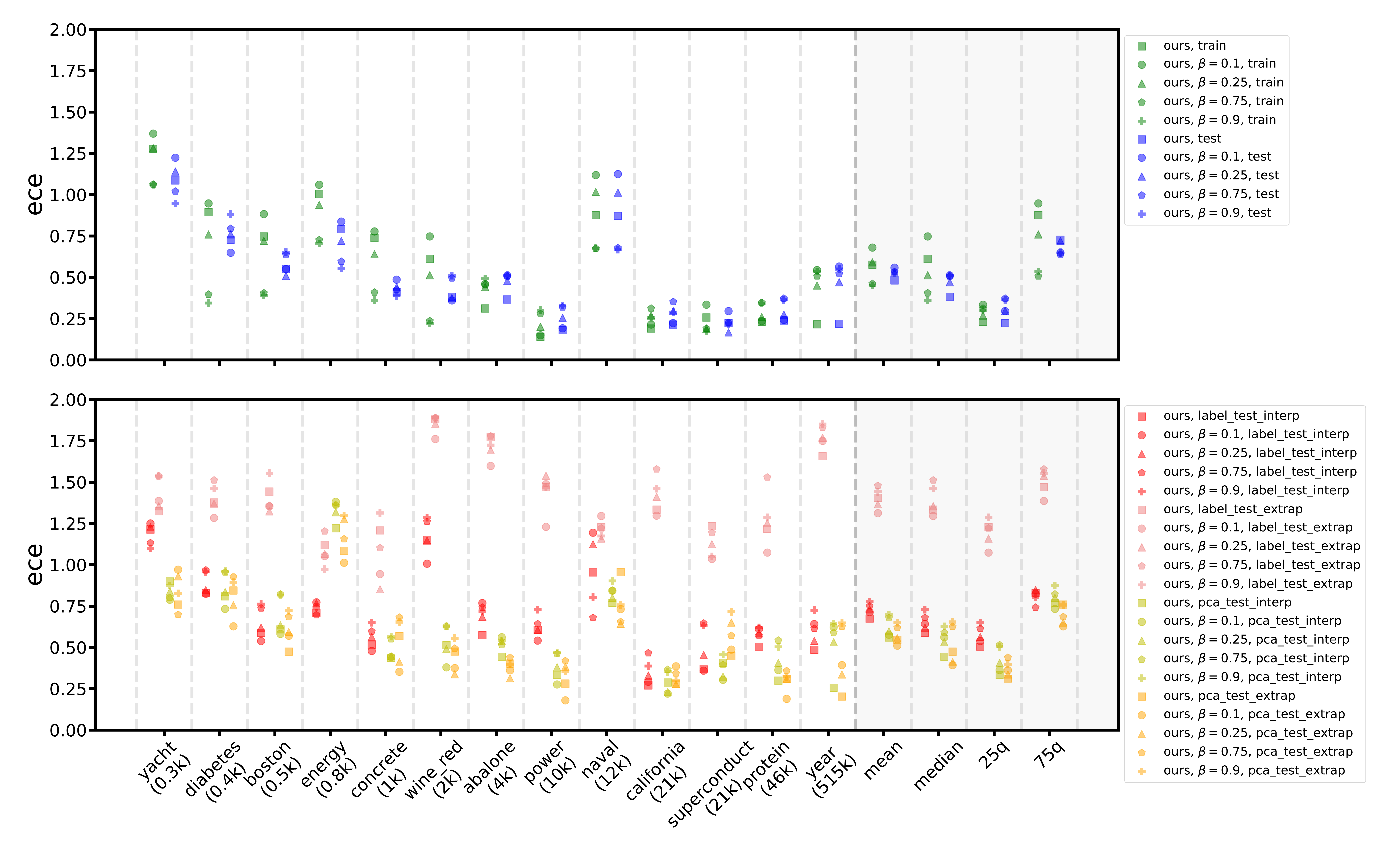}
    \caption{Expected calibration errors (ECEs) for SML-trained networks with hyper-parameters $\beta = 0.1, 0.25, 0.5, 0.75, 0.9$. We consider 13 UCI regression datasets under \iid conditions (top) and under data shift (bottom). $\beta$-values are encoded via plot marker, data splits via color. Each plot point corresponds to a cross-validated trained network. Summarizing statistics (rhs) are indicated by a light grey background.}
    \label{fig:ece_fs}
\end{figure}

\begin{figure}[pbth]
    \includegraphics[width=\textwidth]{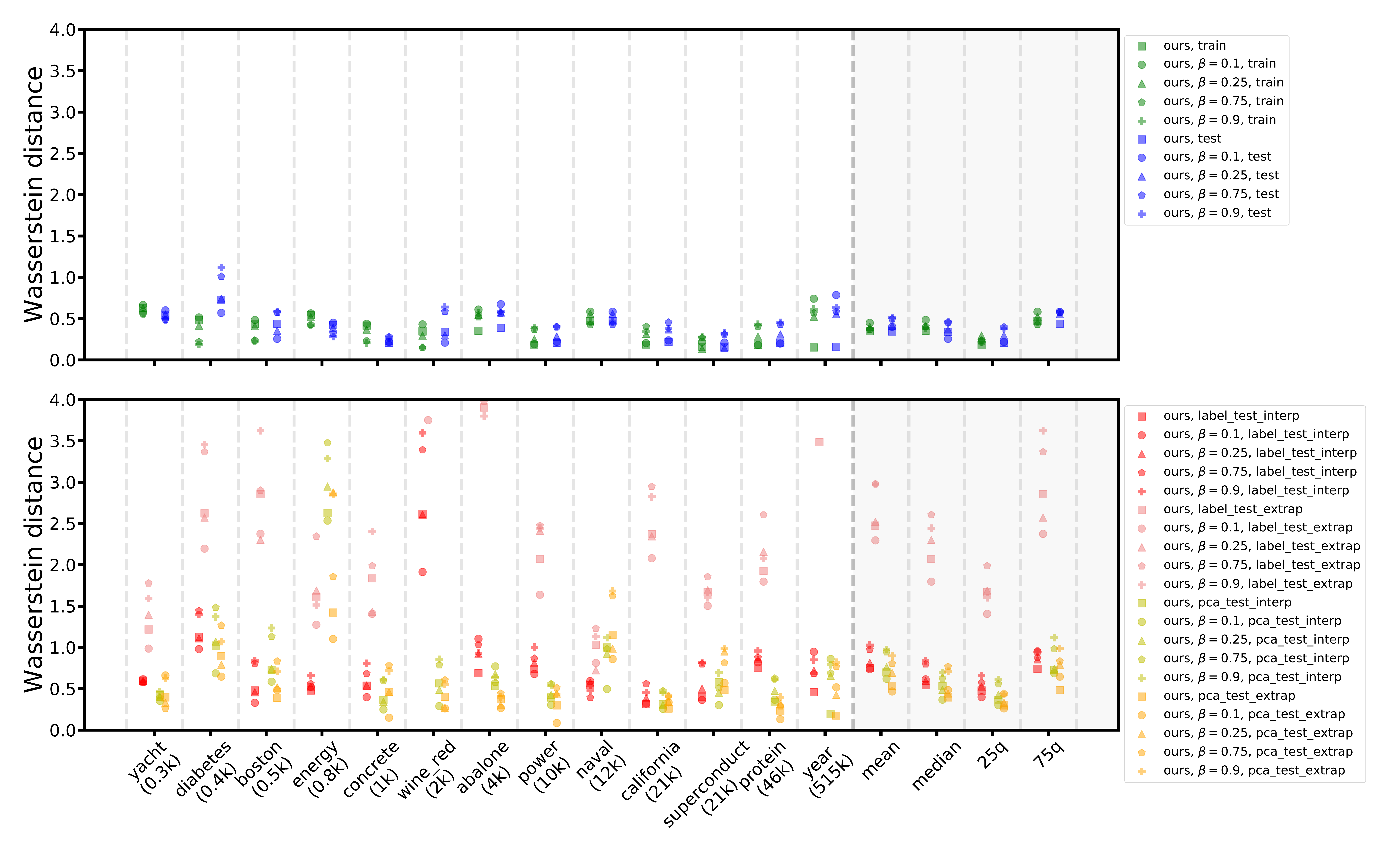}
    \caption{Wasserstein distances for SML-trained networks with hyper-parameters $\beta =\allowbreak 0.1,\allowbreak 0.25, 0.5, 0.75, 0.9$. We consider 13 UCI regression datasets under \iid conditions (top) and under data shift (bottom). $\beta$-values are encoded via plot marker, data splits via color. Each plot point corresponds to a cross-validated trained network. Summarizing statistics (rhs) are indicated by a light grey background.}
    \label{fig:ws_dist_fs}
\end{figure}

Here, we analyze the impact of the SML-parameter $\beta$ on the uncertainty quality of accordingly trained models. For $\beta = 0.1, 0.25, 0.5, 0.75, 0.9$, we observe only relatively small differences in both ECE (see Fig. \ref{fig:ece_fs}) and Wasserstein distance (see Fig. \ref{fig:ws_dist_fs}). $\beta = 0.5$ provides (by a small margin) the best average test set performance in both scores. However, the best-performing $\beta$-value for an individual dataset can vary.

Experiments with $\beta \gg 1$ (not shown here) cause non-convergent training in many cases as primarily uncertainty quality is optimized at the expense of task performance. The opposite extreme case is $\beta = 0$, \ie network optimization without any dropout mechanism. Applying dropout at inference will therefore cause uncontrolled random fluctuations around the network prediction.


\section{In-depth investigation of uncertainty measures}
\label{appendix:unc_measures}

\subsection{Dependencies between uncertainty measures}

\begin{figure}[pbth]
    \centering
    \includegraphics[width=0.8\textwidth]{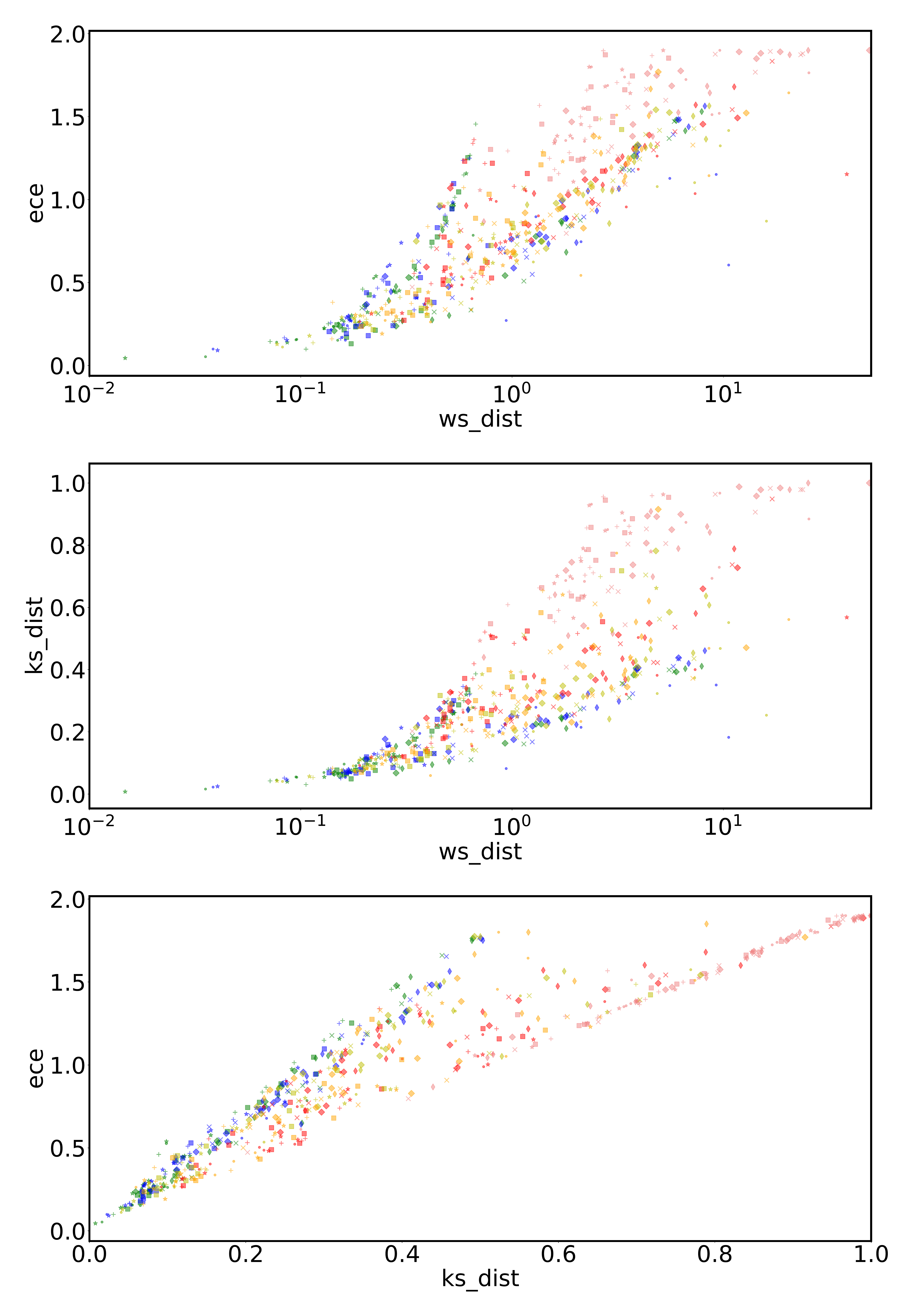}
    \caption{Dependencies between the three uncertainty measures ECE, Wasserstein distance and Kolmogorov-Smirnov distance. Uncertainty methods are encoded via plot markers, data splits via color. Datasets are not encoded and cannot be distinguished (see text for more details). Each plot point corresponds to a cross-validated trained network. The clearly visible deviations from ideal correlations point at the potential of these uncertainty measures to complement one another.} 
    \label{fig:corrs_uncertainty_measures}
\end{figure}

All uncertainty-related measures (NLL, ECE, Wasserstein distance) relate predicted uncertainties to actually occurring model residuals. Each of them putting emphasize on different aspects of the considered samples: NLL is biased towards well-performing models, ECE measures deviations within quantile ranges, Wasserstein distance resolves distances between normalized residuals. The empirically observed dependencies between these uncertainty measures are visualized in Fig.\  \ref{fig:corrs_uncertainty_measures}.
Additionally to Wasserstein distances, we consider Kolmogorov-Smirnov (KS) distances \citep{stephens1974edf} on normalized residuals there.
It estimates a distance between the sample of normalized residuals and a standard Gaussian.
Different from the Wasserstein distance, the KS-distance is not transport-based but determined by the largest distance between the empirical CDFs of the two samples.
It is therefore bounded to $[0,1]$ and unable to resolve differences between samples that strongly deviate from a standard Gaussian one.

While all these scores are expectably correlated, noteworthy deviations from ideal correlation occur.
Therefore, we advocate for uncertainty evaluations based on various measures to avoid overfitting to a specific formalization of uncertainty.  

The data splits in Fig.\ \ref{fig:corrs_uncertainty_measures} are color-coded as follows: train is green, test is blue, pca-interpolate is green-yellow, pca-extrapolate is orange-yellow, label-interpolate is red and label-extrapolate is light red. The mapping between uncertainty methods and plot markers reads: MC is `diamond', MC-LL is `thin diamond', DE is `cross', PU is `point', PU-DE is `pentagon' and \ours is `square'. Some Wasserstein distances lie above the x-axis cut-off and are thus not visualized.

\subsection{Discussion of NLL as a measure of uncertainty}
\label{appendix:nll}

Typically, DNNs using uncertainty are often evaluated in terms of their negative log-likelihood (NLL).
This property is affected not only by the uncertainty, but also by the DNNs performance.
Additionally, it is difficult to interpret, sometimes leading to contraintuitive results, which we want to elaborate on here.
As a first example, take the likelihood of two datasets $x_1=\{0\}$ and $x_2=\{0.5\}$, each consisting of a single point, with respect to a normal distribution $\mathcal{N}(0,1)$.
Naturally, we find $x_1$ to be located at the maximum of the considered normal distribution and deem it the more likely candidate.
But, if we extend these datasets to more than single points, \ie $\tilde x_1= \{0,0.1,0,-0.1,0\}$ and $\tilde x_2=\{0.5,-0.4,0,-1.9,-0.7\}$, it becomes obvious that $\tilde x_2$ is much more likely to follow the intended Gaussian distribution.
Nonetheless, $\text{NLL}(\tilde x_2)\approx 1.4 > 0.9 \approx \text{NLL}(\tilde x_1)$, where
\begin{equation}
    \text{NLL}(y):=\log{\sqrt{2\pi\sigma^2}}+\frac{1}{N}\sum_{i=1}^N \frac{(y_i-\mu)^2}{2\sigma^2}\,.
    \label{eq:appxNLL}
\end{equation}
This may be seen as a direct consequence of the point-wise definition of NLL, which does not consider the distribution of the elements in $\tilde x_i$.
From this observation also follows that a model with high prediction accuracy will have a lower NLL score as a worse performing one if uncertainties are predicted in the same way.
Independent of whether those reflected the ``true'' uncertainty in either case.
This issue can be further substantiated on a second example.
Consider two other datasets $z_1,z_2 $ drawn \iid from Gaussian distributions $\mathcal{N}(0,\sigma_i)$ with two differing values $\sigma_1\!<\!\sigma_2$.
If we determine the NLL of each with respect to its own distribution the offset term in equation (\ref{eq:appxNLL}) leads to $\text{NLL}(z_2)=\text{NLL}(z_1)+\log{(\sigma_2/\sigma_1)}$ with $\log{(\sigma_2/\sigma_1)}>0$.
Although both accurately reflect their own distributions, or uncertainties so to speak, the narrower $z_1$ is more ``likely''.
This offset makes it difficult to assess reported NLL values for systems with heteroskedastic uncertainty.
While smaller is typically ``better'', it is highly data- (and prediction-) dependent which value is good in the sense of a reasonable correlation between performance and uncertainty.


\end{document}